\documentclass[final,3p,times,twocolumn]{elsarticle}

\usepackage{amssymb}

\journal{arXiv}

\usepackage[numbers]{natbib}
\usepackage{amsmath,amssymb}
\usepackage{graphicx}
\usepackage{subcaption}
\usepackage{psfrag}
\usepackage{enumerate}
\usepackage{algorithmic}
\usepackage{algorithm}
\usepackage{acronym}
\usepackage{color, soul}
\usepackage{amsthm}
\usepackage{pifont}
\usepackage{bm}
\usepackage{afterpage}
\usepackage{multicol}
\usepackage{array}
\usepackage[T1]{fontenc}
\usepackage{makecell}
\usepackage{multicol}
\usepackage{cuted}

\usepackage{mathtools,leftindex}
\usepackage{xcolor}
\usepackage{color,soul}
\usepackage{tcolorbox}

\newcommand{\Rz}{\bm{R}_{Z}(\psi)}
\newcommand{\Ry}{\bm{R}_{y^{\prime}}\left(\theta\right)}
\newcommand{\Rx}{\bm{R}_{x^{\prime \prime}}\left( \phi\right)}
\newcommand{\RO}{\leftindex_I^B{\bm{R}}}

\newcommand{\Rone}{\bm{R}_1}
\newcommand{\Rtwo}{\bm{R}_2}
\newcommand{\Rthree}{\bm{R}_3}
\newcommand{\Rfour}{\bm{R}_4}

\makeatletter
\DeclareRobustCommand{\pdot}{\mathbin{\mathpalette\pdot@\relax}}
\newcommand{\pdot@}[2]{%
	\ooalign{%
		$\m@th#1\circ$\cr
		\hidewidth$\m@th#1\cdot$\hidewidth\cr
	}%
}

\begin{document}
	
	\acrodef{UAV}{Unmanned Aerial Vehicle}
	\acrodef{ESC}{Electronic Speed Control}
	\acrodef{BLDC}{Brushless Direct Current}
	\acrodef{PCF}{Power Consumption Factor}
	\acrodef{DoF}{Degrees of Freedom}
	\acrodef{PD}{Proportional Derivative}
	\acrodef{PID}{Proportional Integral Derivative}
	\acrodef{SMC}{Sliding Mode Controller}
	\acrodef{VRS}{Vortex Ring State}
	\acrodef{CoG}{Center of Gravity}
	\acrodef{CAD}{Computer Aided Design}
	\acrodef{CF}{Comparison Factor}
	\acrodef{NPCF}{Nondimensionalized Power Consumption Factor}
	
	\begin{frontmatter}
		
		
\title{Next-Generation Aerial Robots\textemdash Omniorientational Strategies: Dynamic Modeling, Control, and Comparative Analysis}

\author[label1]{Ali Kafili Gavgani\tnoteref{aff1}}
\ead{Ali.Kafili79@sharif.edu; AliKafili@yahoo.com}
\cortext[cor1]{Corresponding Author}

\author[label1]{Amin Talaeizadeh\corref{cor1}}
\ead{amin.talaeizadeh@sharif.edu}

\author[label1]{Aria Alasty}
\ead{aalasti@sharif.edu}

\author[label1]{Hossein Nejat Pishkenari}
\ead{nejat@sharif.edu}

\author[label2]{Esmaeil Najafi\corref{cor1} \tnoteref{aff2}}
\ead{e.najafi@saxion.nl}

\affiliation[label1]{organization={Advanced Research Lab for Control and Agricultural Robotics (Sharif AgRoLab)\\ Department of Mechanical Engineering, Sharif University of Technology},
	city={Tehran},
	country={Iran}}

\affiliation[label2]{organization={Smart Mechatronics and Robotics Research Group\\ Saxion University of Applied Science},
	city={Enschede},
	country={Netherlands}}		



\begin{abstract}
Conventional multi-rotors are under-actuated systems, hindering them from independently controlling attitude from position. In this study, we present several distinct configurations that incorporate additional control inputs for manipulating the angles of the propeller axes. This addresses the mentioned limitations, making the systems "omniorientational". We comprehensively derived detailed dynamic models for all introduced configurations and validated by a methodology using Simscape Multibody simulations.
Two controllers are designed: a sliding mode controller for robust handling of disturbances and a novel PID-based controller with gravity compensation integrating linear and non-linear allocators, designed for computational efficiency. A custom control allocation strategy is implemented to manage the input-non-affine nature of these systems, seeking to maximize battery life by minimizing the "Power Consumption Factor" defined in this study. Moreover, the controllers effectively managed harsh disturbances and uncertainties.
Simulations compare and analyze the proposed configurations and controllers, majorly considering their power consumption. Furthermore, we conduct a qualitative comparison to evaluate the impact of different types of uncertainties on the control system, highlighting areas for potential model or hardware improvements. The analysis in this study provides a roadmap for future researchers to design omniorientational drones based on their design objectives, offering practical insights into configuration selection and controller design. 
This research aligns with the project SAC-1, one of the objectives of  Sharif AgRoLab. 

%

			
		\end{abstract}
		\acresetall

		
	\end{frontmatter}
	
	
	\section{Introduction}
	
	\label{intro}
	Multi-rotors have become highly popular due to their versatility and agility, finding use in a wide range of applications. These include package transportation~\cite{package, PackDel, packdelEurop}, search and rescue operations~\cite{search,searchRescueFranc,searchHel,searchHel2}, agriculture~\cite{agriculture, Agriculture2, shahrouz, agriHel,agriHel2,agriHel3,agriHel4}, non-destructive testing~\cite{nonedestructive,NDTRA1}, aerial surveillance and remote sensing~\cite{surveillance, remoteHel, remoteHel2,remoteHel3}.
	
	Conventional multi-rotors (e.g., a quadrotor) are under-actuated systems, hindering them from performing complex maneuvers and independently controlling of attitude from position.
To overcome these limitations, \citet{ryll} and \cite{strangemechanism} proposed incorporating additional independent actuators to make their drone fully-actuated. Both studies utilized feedback linearization control techniques based on third-order differential equations of motion.

\citet{zheng} and \citet{badr} achieved full actuation in their systems using thrust-vectoring mechanisms, enabling the drone to hover with an inclined body.	
	\citet{allenspach} and \citet{voliro} constructed hexacopters with tilting rotors, capable of tracking all six \ac{DoF} independently. \citet{voliro} incorporated a PID controller to determine the desired force vector and Two cascade controllers to generate the desired moment vector.
	\citet{santosfastallocation} introduced a fast control allocation method for an over-actuated quadrotor to reduce real-time computation costs. A \ac{PD} \ac{SMC} is proposed and simulated by \citet{PDSliding} to control an over-actuated quadrotor, integrating a genetic algorithm to optimally tune the controller parameters. A different over-actuated configuration is proposed by \citet{KafiliOman}, introducing a \ac{SMC}-based controller incorporating a none-linear mapping in control inputs to deal with control allocation problems. 
 
	\citet{tala-vortex} introduced a yaw-rate control to mitigate the effects of \ac{VRS}.
	\citet{zargar} developed a quadrotor with rotor axes tilted inwards, taking a dihedral angle to explore a different approach to overcome \ac{VRS}. 	
	\sethlcolor{yellow!40}
	Some studies have assessed the different applications of sliding mode control~\cite{SMC1rev1,SMC2rev1,SMC4rev2}.

	
	

Although previous works have made strides in thrust-vectoring and full actuation, there remains a gap in the literature on comprehensive comparisons of omniorientational configurations and controllers. Such comparisons would allow for a nuanced understanding of how different configurations and control strategies affect energy consumption and performance across the full six-DoF.

\begin{figure}[ht]
 	\centering
 	\begin{subfigure}{0.4\linewidth}
 		\includegraphics[width=3.5cm]{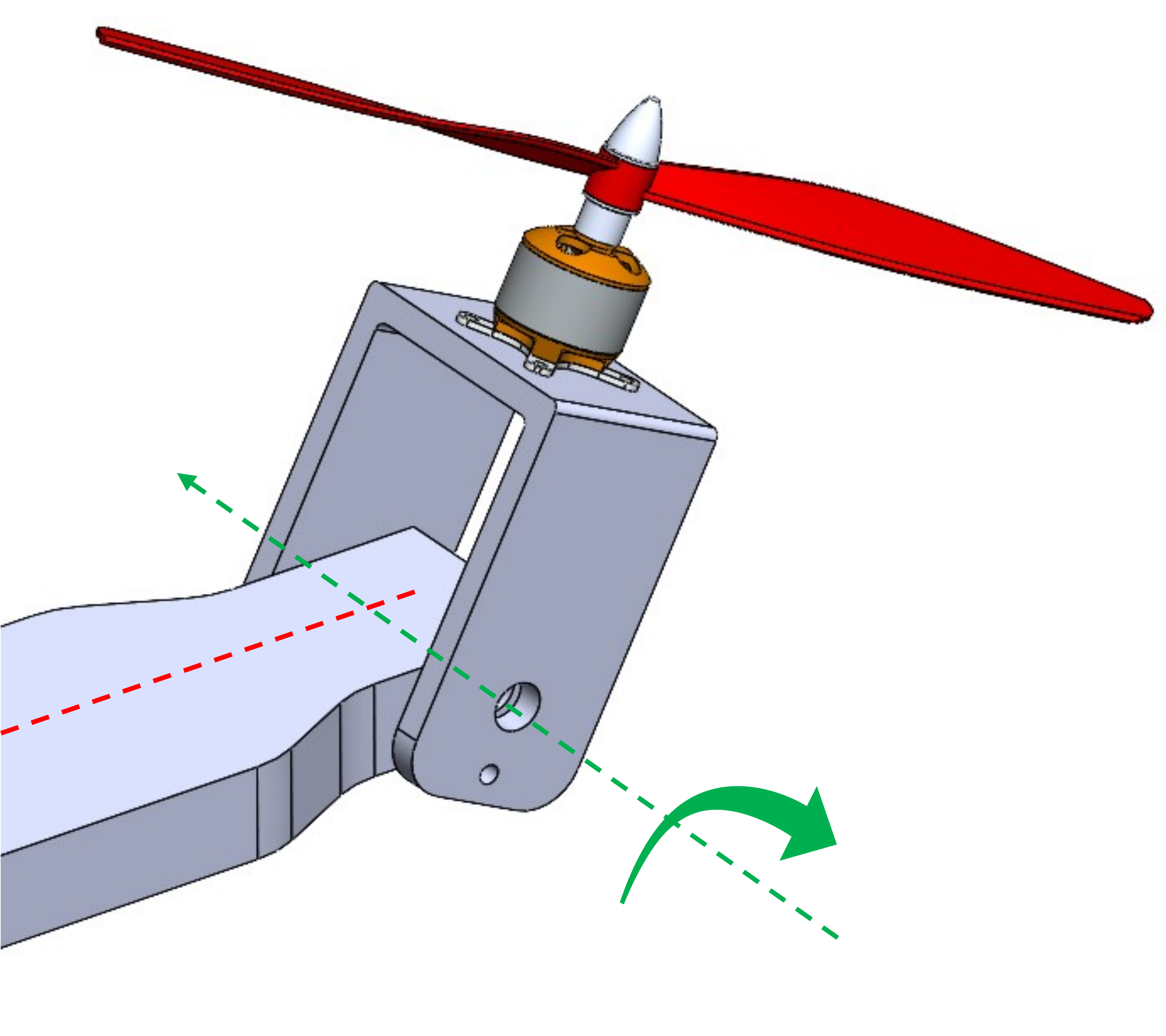}
 		\caption{hedral rotation}
 	\end{subfigure}
 	\hspace{0.5cm}
 	\begin{subfigure}{0.4\linewidth}
 		\includegraphics[width=3.5cm]{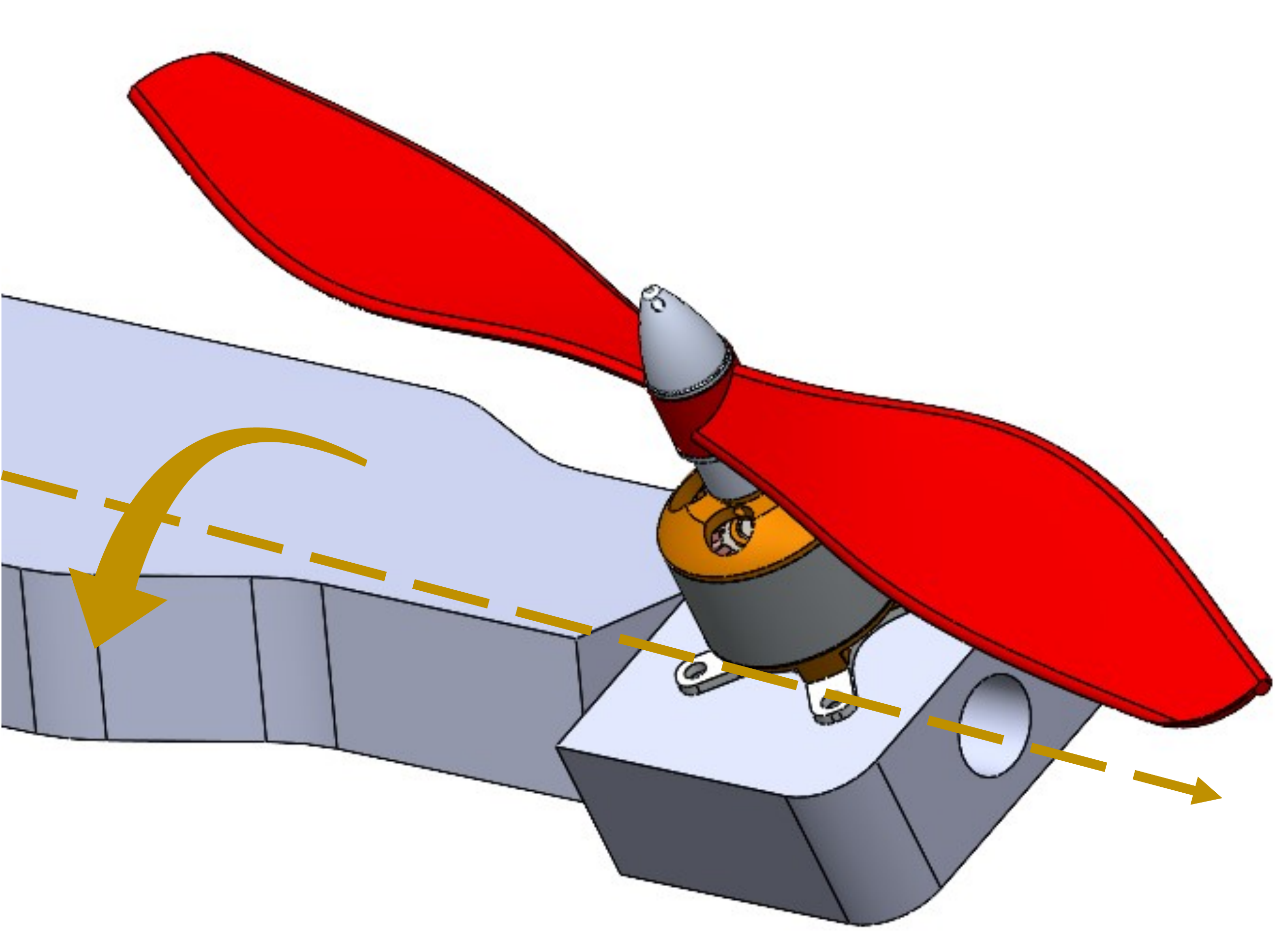}
 		\caption{tilt rotation}
 	\end{subfigure}
 	\caption{Illustration of two types of propeller axis rotations, referred to in this study as \textit{hedral rotation} and \textit{tilt rotation}.}
 	\label{fig:hedral-vs-tilt}
 \end{figure}

 We represent four unique drone configurations featuring rotatable propeller axes. Our target systems are "omniorientational", indicating their ability to independently control position and attitude across all six \ac{DoF}. The terms "hedral rotation" and "tilt rotation" are used in this article to refer to two distinct methods of axis rotation. 	Refer to Figure~\ref{fig:hedral-vs-tilt} for a graphical representation of hedral and tilt rotations. Detailed comparative analysis examines energy efficiency, robustness, and control efficacy across configurations and controllers.
 
\begin{figure*}[t]
 	
 	\begin{subfigure}{0.4\linewidth}
 		\includegraphics[height=5cm]{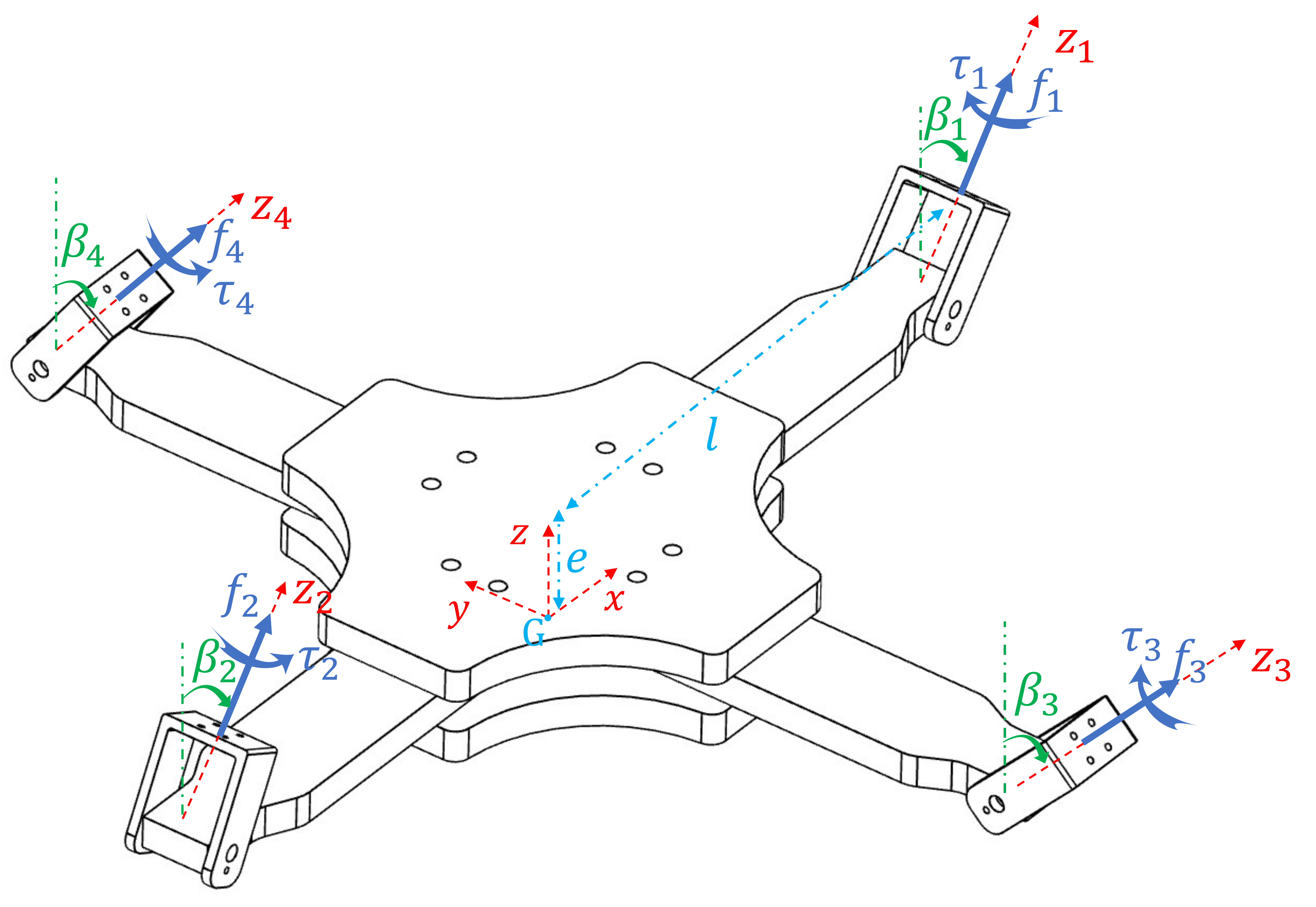}
 		\caption{The "Hedral" configuration with four servo motors to manipulate the hedral angle of each BLDC motor independently.}
 	\end{subfigure}
 	\hspace{2cm}
 	\begin{subfigure}{0.4\linewidth}
 		\centering
 		\includegraphics[height=5cm]{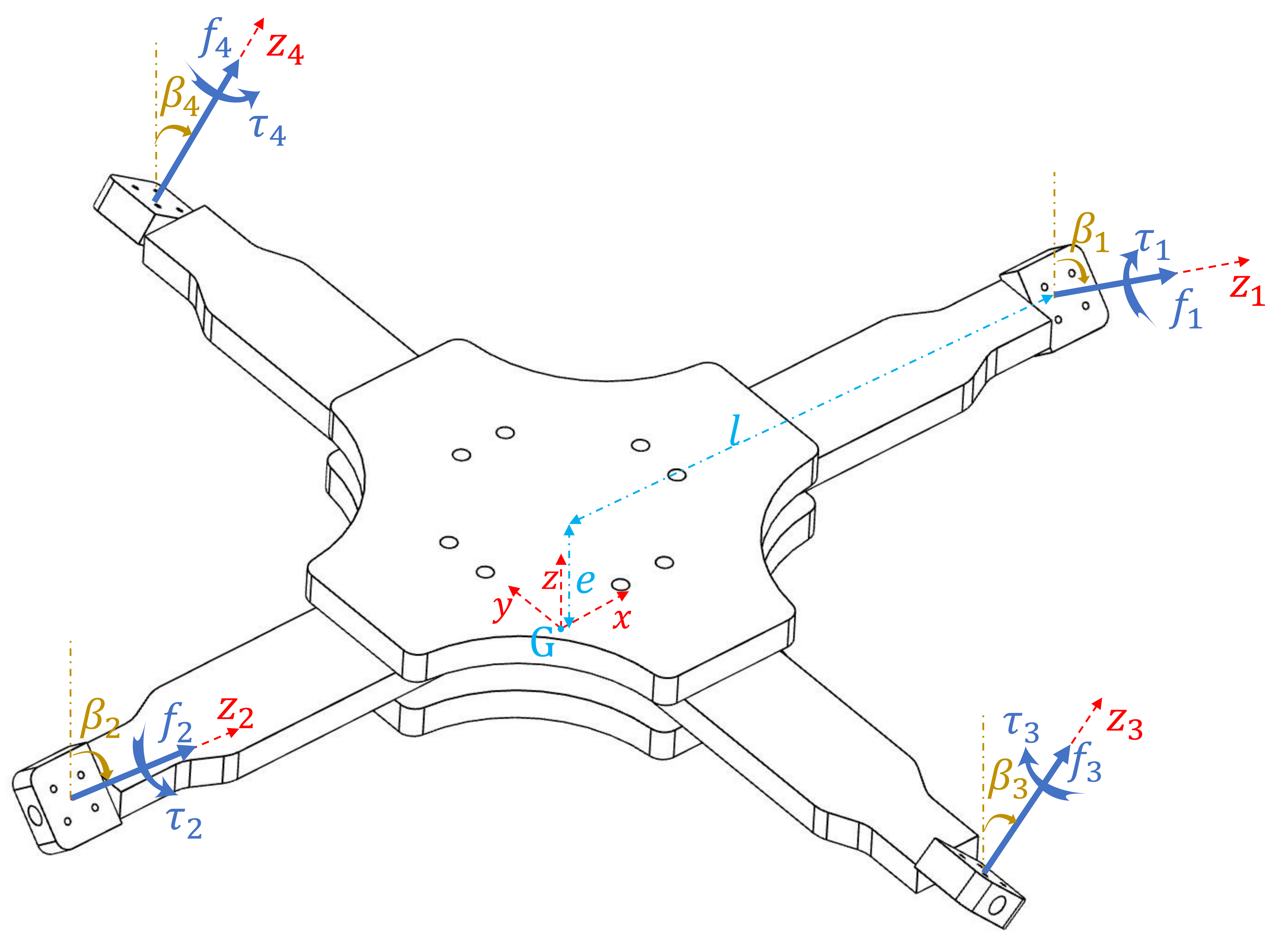}
 		\caption{The "Tilt" configuration with four servo motors to manipulate the tilt angle of each BLDC motor independently.}
 	\end{subfigure}
 	
 	\vspace{2ex}
 	
 	\begin{subfigure}{0.45\linewidth}
 		\includegraphics[height=5cm]{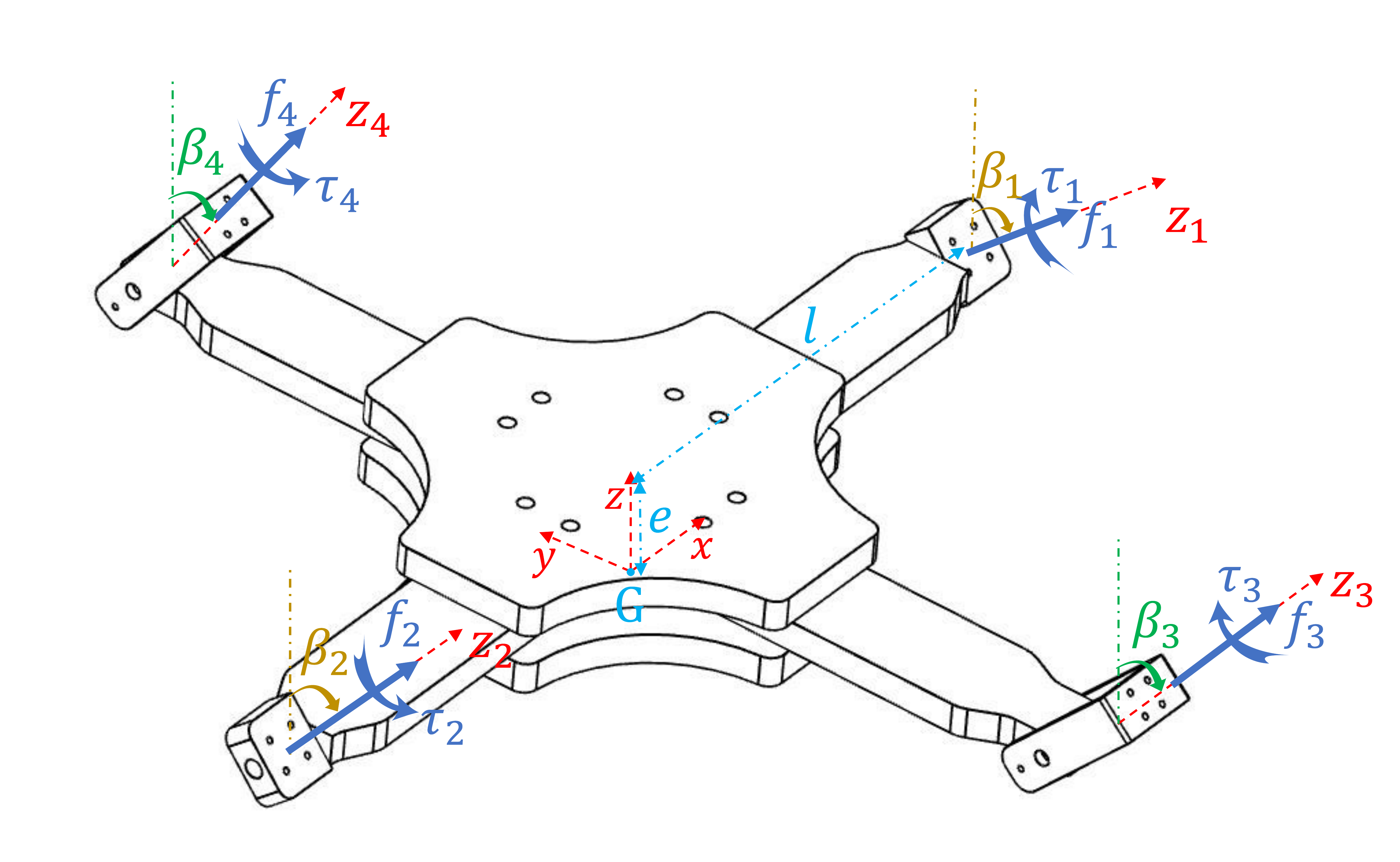}
 		\caption{The "Half-Tilt" configuration employing two servo motors to control the hedral angle of two BLDC motors, and two additional servo motors for manipulating the tilt angle of the remaining two BLDC motors.}
 	\end{subfigure}
 	\hspace{0.5cm} 
 	\begin{subfigure}{0.5\linewidth}
 		\centering
 		\includegraphics[height=4.5cm]{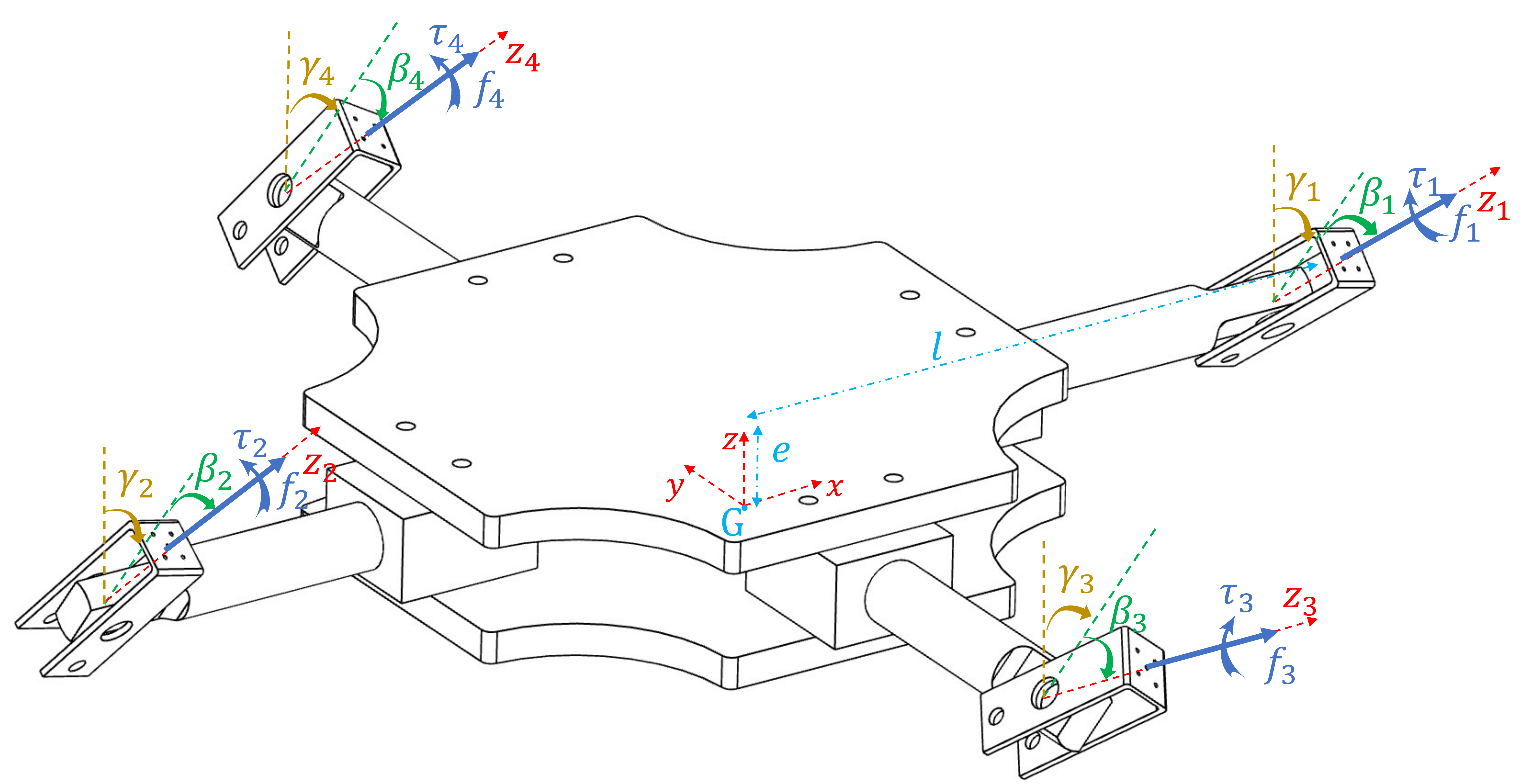}
 		\caption{The "Tilt-Hedral" configuration integrating eight servo motors to manipulate both hedral and tilt angles of each BLDC motor independently.}
 	\end{subfigure}
 	
 	\caption{Different configurations with their free body diagram and coordinate systems.}
 	\label{fig:configs}
 \end{figure*}

\begin{figure}
 	\centering
 	\includegraphics[width=\linewidth]{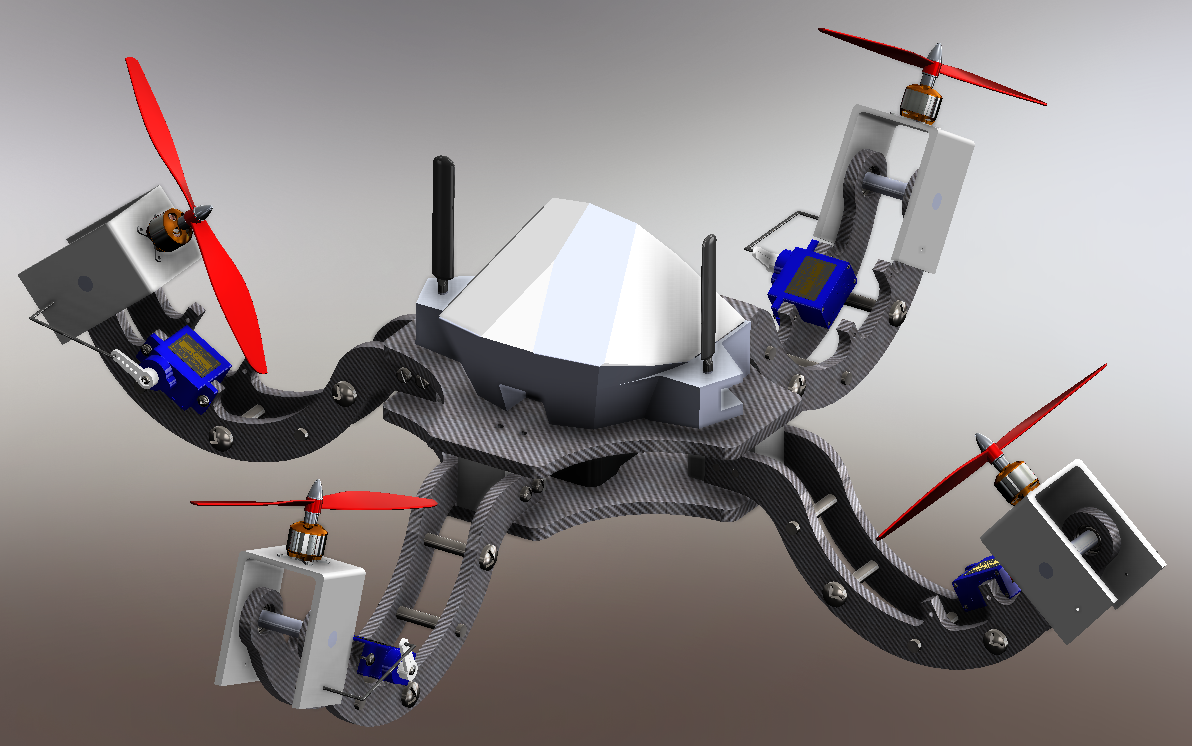}
 	\caption{A conceptual design for the \textit{Hedral} configuration.}
 	\label{fig:conceptual}
 \end{figure}

 The simplified visualization of proposed configurations are represented in Figure~\ref{fig:configs} emphasizing on the physical parameters, notations and coordinate systems. They involve four \ac{BLDC} motors with attached propellers along with integrated servo motors for manipulating hedral and/or tilt angles. A conceptual design for the \textit{Hedral} configuration is illustrated in Figure~\ref{fig:conceptual}.

 The dynamic equations of motion for each configuration are derived comprehensively using the Newton-Euler method, validated by applying our model-based controller to CAD models in MATLAB Simscape Multibody environment.
 
 We have developed two effective controllers: a sliding mode controller for robust handling of disturbances and a novel PID-based controller with gravity compensation integrating linear and non-linear allocators, designed for computational efficiency. A custom control allocation strategy is implemented to manage the input-non-affine nature of these systems, seeking to maximize battery life by minimizing the \ac{PCF} and \ac{NPCF} defined in this study.
 
 A comparison framework assessing power consumption and robustness under uncertainties and disturbances is developed in this study. This framework introduces a power factor and an Uncertainty Impact Factor (UIF) for evaluating configurations and controllers, highlighting areas for potential model or hardware improvements, and offering insights into each configuration’s practical capabilities and operational strengths.

 The main contribution of this paper is exploring a comprehensive assessment of various possible configurations for omniorientational drones \textemdash systems designed to overcome the limitations of conventional multi-rotors by enabling independent control across all \ac{DoF}. This enables a comparative analysis of the different configurations and controllers. 
 The unique analysis in this paper provides a roadmap for future researchers to design omniorientational drones based on their design objectives, offering practical insights into configuration selection and controller design.

	
	
	

	
	This research is in alignment with the project SAC-1, one of the primary objectives of the Laboratory of \textit{Advanced Control Systems and Agricultural Robotics of Sharif University of Technology (Sharif AgRoLab)}, which is the development of the \textit{Sharif Agrocopter}. This specialized drone is designed to navigate with ease and precision through dense garden environments abundant with trees and foliage. Its primary functions encompass conducting essential investigations, including the application of artificial intelligence for pest detection and management.

	\section{Dynamic Modeling} \label{sec:dynamics}
The dynamic equations of four distinct omniorientational drone configurations, with the simplified visualizations depicted in Figure \ref{fig:configs}, are derived in this section.
In Figure \ref{fig:configs}, \textit{l} denotes the distance between the point of application of each propeller's thrust force and the \ac{CoG} along the arms, while \textit{e} represents the eccentricity of the \ac{CoG} relative to this point, perpendicular to the arms' axis.
 Minor alternations in \textit{l}, \textit{e}, and inertia moments due to the rotation of the propeller axis are also neglected.




\subsection{Coordinate Systems and Rotation Matrices} 
The inertial coordinate system is denoted by $XYZ$, while \textit{xyz} refers to the body-fixed frame attached to the \ac{CoG}. The frame $x_{n} y_{n} z_{n}$ is attached to the $n$-th propeller. The angles $\beta$ and $\gamma$ indicate the rotation angles of the propeller axes. Refer to Fig.~\ref{fig:configs} for a clearer understanding.


	The sequence outlined in (\ref{XYZtoxyz}) transforms the $XYZ$ frame into the \(xyz\) frame \cite{ginsberg}.
	\begin{equation}
		\label{XYZtoxyz}
		X Y Z \stackrel{\Rz}{\longrightarrow} x^{\prime} y^{\prime} z^{\prime} \stackrel{\Ry}{\longrightarrow} x^{\prime \prime} y^{\prime \prime} z^{\prime \prime} \stackrel{\Rx}{\longrightarrow} x y z
	\end{equation}
	
	Thus, the inertial-to-body rotation matrix is obtained as
	\begin{equation}
		\label{R0}
		\RO= \Rx\Ry\Rz , 
	\end{equation}
	where $I$ index indicates the inertia frame and $B$ represents the body frame.
	
\begin{figure}[t]
		\centering
		\includegraphics[width=\linewidth]{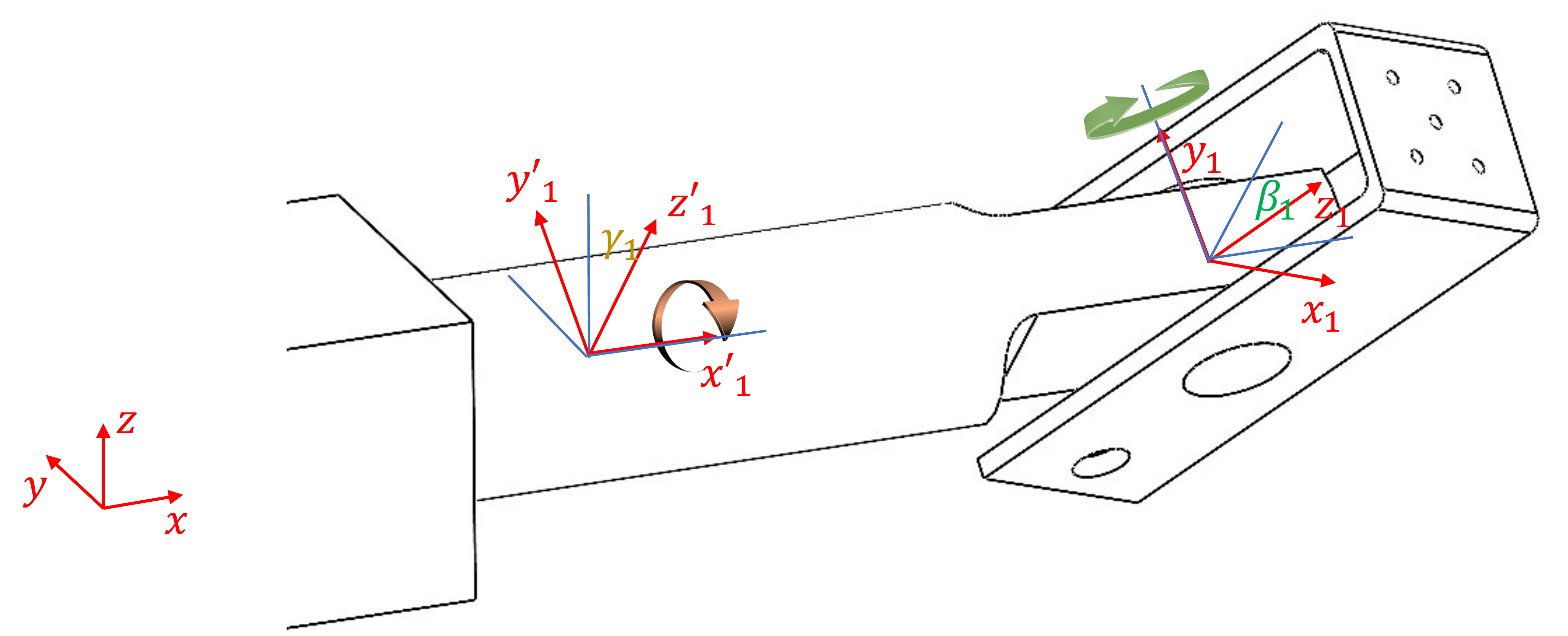}
		\caption{Detailed demonstration of rotations in an arm of the Tilt-Hedral configuration.}
		\label{fig:tilt-hedral-arm}
\end{figure}

	$\bm{R_n}$, $n=1,2,3,4$ is unique for each configuration, enabling the transformation from the $xyz$ frame to the specific $x_{n} y_{n} z_{n}$ frame of that configuration. This approach allows for a unified representation of the dynamic equations, eliminating the need to write separate series of dynamics for each configuration. in the following, consider employing corresponding $\bm{R}_n$ matrices to derive the dynamics for the desired configuration.
	
	a) For the "Hedral" configuration
	\begin{equation}
		\begin{aligned}
			\Rone&=\bm{R}_y(\beta_1),\quad   
			\Rtwo=\bm{R}_y(\beta_2),\\
			\Rthree&=\bm{R}_x(\beta_3), \quad
			\Rfour=\bm{R}_x(\beta_4).
		\end{aligned}   
	\end{equation}
	
	b) For the "Tilt" configuration
	\begin{equation}
		\begin{aligned}
			\Rone&=\bm{R}_x(\beta_1), \quad   
			\Rtwo=\bm{R}_x(\beta_2),\\   
			\Rthree&=\bm{R}_y(\beta_3), \quad   
			\Rfour=\bm{R}_y(\beta_4).
		\end{aligned}   
	\end{equation}
	
	c) For the "Half-Tilt" configuration
	
	\begin{equation}
		\begin{aligned}
			\Rone&=\bm{R}_x(\beta_1),\quad   
			\Rtwo=\bm{R}_x(\beta_2),\\
			\Rthree&=\bm{R}_x(\beta_3), \quad
			\Rfour=\bm{R}_x(\beta_4).
		\end{aligned}   
	\end{equation}
	
	d) For the "Tilt-Hedral" configuration:
	This configuration merits closer discussion. Thus, one of the arms is specifically explored.
	As illustrated in Fig. \ref{fig:tilt-hedral-arm}, the $xyz$ frame undergoes a rotation around the $x$-axis by an angle of $\gamma_1$, resulting in the $x^{\prime}_1y^{\prime}_1z^{\prime}_1$ frame.
	The subsequent rotation around the $y^{\prime}_1$ axis by an angle of $\beta_1$ leads to the formation of the $x_1y_1z_1$ frame.
	Through this successive tilt and hedral rotations, the $xyz$ frame transforms into the $x_1y_1z_1$.
	
	\begin{equation}
		\begin{aligned}
			\Rone&=\bm{R}_{y^\prime_1}(\beta_1)\bm{R}_x(\gamma_1),\quad   
			\Rtwo=\bm{R}_{y^\prime_2}(\beta_2)\bm{R}_x(\gamma_2),\\
			\Rthree&=\bm{R}_{x^\prime_3}(\beta_3)\bm{R}_y(\gamma_3), \quad
			\Rfour=\bm{R}_{x^\prime_4}(\beta_4)\bm{R}_y(\gamma_4).
		\end{aligned}   
	\end{equation}
	

	\subsection{Aerodynamics}	
	The thrust force and drag torque produced by a motor-propeller pair are directly proportional to the square of its angular velocity. Thus, the thrust force generated by the $n$-th propeller is given by
	\begin{equation}
		f_n = k_f \omega_n^2,
	\end{equation}
	while the corresponding drag torque is expressed as 
	\begin{equation}
		\tau_n = k_m \omega_n^2,
	\end{equation}
	for $n = 1, 2, 3, 4$. Here, $k_f$ and $k_m$ represent the aerodynamic coefficients of the propellers, and $\omega_n$ is the angular velocity of the $n$-th propeller.
	
	As illustrated in Fig.~\ref{fig:configs}, $f_n$ and $\tau_n$ are generated along the $z_n$ axis of the frame $x_n y_n z_n$. Therefore, the thrust and drag torque vectors of the $n$-th propeller are given by
	\begin{align}
		\bm{f_n} &= f_n \bm{e}_{z_n}, \label{thrust_prop} \\
		\bm{\tau_n} &= \tau_n \bm{e}_{z_n} \label{torque_prop},
	\end{align}
	where $\bm{e_{z_n}}$ represents the unit vector along the $z_n$ axis.
	
\subsection{Dynamics of the Tilt/Hedral Angle Manipulation System}
The motors responsible for manipulating the tilt/hedral angles can be directly attached to the main rotor or connected via a linkage. This system is modeled using first-order dynamics as follows \cite{allenspach}:
\begin{equation}
	\dot{b}_n = \frac{1}{T_n} (b_{\text{ref},n} - b_n),
\end{equation}
where $T_n$ is the time constant of the $n$-th manipulation system and $b_{\text{ref},n}$ indicates the $n$-th tilt/hedral angle desired by the control system.

However, given the fast response of the actuators, their transient dynamics can be disregarded.
	
	\subsection{Derivation of Translational Equations of Motion}
	The position vector of \ac{CoG} is defined as
%
	\begin{equation}
		\leftindex^I{\bm{p}} = \left[ \begin{array}{ccc}
			X & Y & Z \end{array} \right]^\intercal .
	\end{equation}
	
	The resultant of thrust forces is represented by
	\begin{equation}
		\label{f tot}
		\bm{f}=\bm{f_1}+\bm{f_2}+\bm{f_3}+\bm{f_4}.
	\end{equation}
	
	Considering (\ref{thrust_prop}), $\bm{f}$ would be represented in the inertia frame by
	\begin{equation}
		\begin{aligned}
			\leftindex^I{\bm{f}}=& 
			\RO^\intercal \Rone^\intercal \left[\begin{array}{lll}
				0 & 0 & f_1
			\end{array}\right]^\intercal +
			\RO^\intercal \Rtwo^\intercal
			\left[\begin{array}{lll} 
				0 & 0 & f_2
			\end{array}\right]^\intercal  \\
			+	&	\RO^\intercal \Rthree^\intercal \left[\begin{array}{lll}
				0 & 0 & f_3
			\end{array}\right]^\intercal + 
			\RO^\intercal \Rfour^\intercal
			\left[\begin{array}{lll}
				0 & 0 & f_4
			\end{array}\right]^\intercal .
		\end{aligned}
	\end{equation}
	
	Consider (\ref{eq:f=ma}) describing the transnational motion
	\begin{equation}
		\label{eq:f=ma}
		\leftindex^I{\bm{f}}+m \leftindex^I{\bm{g}}=m \leftindex^I{\bm{\ddot{p}}},
	\end{equation}
	where $\bm{g}$ indicates the gravitational acceleration vector.
	Having $\bm{\ddot{p}}=\left[\begin{array}{lll}
		\ddot{X} & \ddot{Y} &  \ddot{Z}
	\end{array}\right]^\intercal$, the dynamic equations for translation is calculated by
	\begin{equation}
		\label{positioneq}
		\begin{aligned}
			 \bm{\ddot{p}}=&
			\RO^\intercal \Rone^\intercal \left[\begin{array}{c}
				0 \\ 0 \\ f_1/m
			\end{array}\right]  
			 + 	\RO^\intercal \Rtwo^\intercal
			\left[\begin{array}{c} 
				0 \\ 0 \\ f_2/m
			\end{array}\right] \\
			+&\RO^\intercal \Rthree^\intercal \left[\begin{array}{c}
				0 \\ 0 \\ f_3/m
			\end{array}\right] 
			 +	\RO^\intercal \Rfour^\intercal
			\left[\begin{array}{c}
				0 \\ 0 \\ f_4/m
			\end{array}\right] 
			+ \left[\begin{array}{c}
				0 \\ 0 \\ -g
			\end{array}\right] .
		\end{aligned}
	\end{equation}

	
	\subsection{Derivation of Rotational Equations of Motion}
	Angular velocity of the frame $xyz$ is
	\begin{equation}
		\bm{\omega}=\dot{\phi} \bm{e}_{x^{\prime\prime}}+ \dot{\theta} \bm{e}_{y^{\prime}}+ \dot{\psi} \bm{e}_{Z} .
	\end{equation}
	
	Expressing $\bm{\omega} $ in the body frame results in
	

\begin{equation}
	\label{omegaxyz}
	\begin{aligned}				
		\leftindex^B{\bm{\omega}} =&  
		\Rx \left[\begin{array}{c}
			\dot{\phi} \\ 0 \\ 0
		\end{array}\right]  
		+ \Rx \Ry \left[\begin{array}{c}
			0 \\ \dot{\theta} \\ 0
		\end{array}\right]  
		+ \RO \left[\begin{array}{c} 
			0 \\ 0 \\ \dot{\psi}
		\end{array}\right]  .
	\end{aligned}
\end{equation}
	
	Arrays by $\phi$, $\theta$, $\psi$ and their time derivatives are defined as
	\begin{equation}
		\begin{aligned}
		\bm{o}=& \left\{ \begin{array}{ccc}
			\phi & \theta & \psi \end{array} \right \}^\intercal , \\
		\dot{\bm{o}}=&\left\{ \begin{array}{ccc}
		\dot{\phi} & \dot{\theta} & \dot{\psi} \end{array} \right\}^\intercal , \\
	\ddot{\bm{o}}=&\left\{ \begin{array}{ccc}
		\ddot{\phi} & \ddot{\theta} & \ddot{\psi} \end{array} \right\}^\intercal .			
		\end{aligned}
	\end{equation}
	 
	Angular acceleration of the frame $xyz$ is calculated by
	 \begin{equation}
	 	\bm{\alpha}=\bm{\dot{\omega}}.
	 \end{equation}
	 
	 If all the elements of $\bm{\omega}$ is expressed in body frame, i.e. $\leftindex^B{\bm{\omega}}$, the angular acceleration can be calculated by
	 \begin{equation}
	 	\label{alpha}
	 	\leftindex^B{\bm{\alpha}}=\frac{\partial \leftindex^B{\bm{\omega}}}{\partial \bm{o}} {\bm{\dot{o}}} + \frac{\partial \leftindex^B{\bm{\omega}}}{\partial \bm{\dot{o}}} {\bm{\ddot{o}}} .
	 \end{equation}
	
	Consider (\ref{sigmaMH}) describing the rotational motion \cite{ginsberg},
	
	\begin{equation}
		\label{sigmaMH}
		\sum_{n=1}^4 \leftindex^B{\bm{r}}_{n / G} \times \leftindex^B{\bm{f}}_n+\sum_{n=1}^4 \leftindex^B{\bm{\tau}}_n= \bm{J}_B \leftindex^B{\bm{\alpha}} + \leftindex^B{\bm{\Omega}} \bm{J}_B \leftindex^B{\bm{\omega}} ,
	\end{equation}
	%
%
	 were $\bm{J}_B$ is the mass moment of inertia of the body expressed in the body frame. $\leftindex^B{\bm{\Omega}}$ indicates the skew-symmetric matrix of $\leftindex^B{\bm{\omega}}$. The position vector from \ac{CoG} to the propeller $n$ is represented by  $\bm{r}_{n / G}$.

	%
	
Considering (\ref{thrust_prop}) and (\ref{torque_prop}) the extended formulation of $ \sum_{n=1}^4 \leftindex^B{\bm{r}}_{n / G} \times \leftindex^B{\bm{f}}_n$ is computed using the equation

	\begin{equation}
		\label{moment tot}
	\begin{aligned}
		&\sum_{n=1}^4 \leftindex^B{\bm{r}}_{n / G} \times \leftindex^B{\bm{f}}_n = \\
		&\left[\begin{array}{c}
			+l \\ 0 \\ e
		\end{array}\right] \times \Rone \left[\begin{array}{c}
			0 \\ 0 \\ f_1
		\end{array}\right]   
		+\left[\begin{array}{c}
			-l \\ 0 \\ e
		\end{array}\right]  \times \Rtwo \left[\begin{array}{c}
			0 \\ 0 \\ f_2
		\end{array}\right]   \\
		+&\left[\begin{array}{c}
			0 \\ -l \\ e
		\end{array}\right]  \times \Rthree \left[\begin{array}{c}
			0 \\ 0 \\ f_3
		\end{array}\right]   
		+\left[\begin{array}{c}
			0 \\ +l \\ e
		\end{array}\right]  \times \Rfour \left[\begin{array}{c}
			0 \\ 0 \\ f_4
		\end{array}\right]  . 
	\end{aligned}
\end{equation}
	
	Also,
	
\begin{equation}
	\label{tau tot}
	\begin{aligned}
		& \sum_{n=1}^4  \leftindex^B{\bm{\tau}}_n
		= \\
		&\Rone \left[\begin{array}{c}
			0 \\ 0 \\ -\tau_1
		\end{array}\right]  + 
		\Rtwo \left[\begin{array}{c}
			0 \\ 0 \\ \tau_2
		\end{array}\right]  
		+ \Rthree \left[\begin{array}{c}
			0 \\ 0 \\ -\tau_3
		\end{array}\right]  
		+\Rfour \left[\begin{array}{c}
			0 \\ 0 \\ \tau_4
		\end{array}\right]   . 
		&
	\end{aligned}
\end{equation}	
	
Considering the control strategy discussed in the next section, the dynamic equations for the orientation of body are needed in the form of $\ddot{\phi}$, $\ddot{\theta}$, and $\ddot{\psi}$. By substituting (\ref{alpha}) into (\ref{sigmaMH}), we obtain the desired equations, as represented by (\ref{orient dyn}). 	


	

\begin{equation}
			\label{orient dyn}
	\begin{aligned}
		\bm{\ddot{o}} &=\left( \frac{\partial \leftindex^B{\bm{\omega}}}{\partial \bm{\dot{o}}} \right)^{-1}
		\Bigg(
		\bm{J}^{-1} \bigg( \sum_{n=1}^4 \leftindex^B{\bm{r}}_{n / G} \times \leftindex^B{\bm{f}}_n+\sum_{n=1}^4 \leftindex^B{\bm{\tau}}_n
		\bigg) \\
		&  - \bm{J}^{-1}\leftindex^B{\bm{\Omega}} \bm{J}_B \leftindex^B{\bm{\omega}}
		-  \frac{\partial \leftindex^B{\bm{\omega}}}{\partial \bm{o}} {\bm{\dot{o}}}
		\Bigg).
	\end{aligned}
\end{equation}

In conclusion, with the generalized coordinates array given by
\begin{equation}
	\bm{\bar{x}}= \begin{array}{cccccc}
		\{ X & Y & Z & \phi & \theta & \psi \}^\intercal \end{array} ,
\end{equation}
and the arrays containing the time derivatives of its elements expressed as
\begin{equation}
	\dot{\bm{\bar{x}}}= \begin{array}{cccccc}
		\{ \dot{X} & \dot{Y} & \dot{Z} & \dot{\phi} & \dot{\theta} & \dot{\psi}\}^\intercal \end{array} ,
\end{equation}
\begin{equation}
	\ddot{\bm{\bar{x}}}= \begin{array}{cccccc}
		\{\ddot{X} & \ddot{Y} & \ddot{Z} & \ddot{\phi} & \ddot{\theta} & \ddot{\psi}\}^\intercal \end{array} ,
\end{equation}
Equations (\ref{positioneq}) and (\ref{orient dyn}) yield the dynamic equations of motion in the form
\begin{equation}
	\label{dyn}
	 \ddot{{\bar{\bm{x}}}} =\left[ \begin{array}{cc}
		 	\ddot{\bm{p}} \\
		 	\ddot{\bm{o}}
		 \end{array} \right].
\end{equation}

 As mentioned before, consider using $\Rone, \Rtwo, \Rthree, \Rfour$ corresponding to the related configuration. The correctness of these equations is verified by the method discussed in Subsection \ref{subsec:verify}.

	\section{Controller Design}
	In this section, a \ac{SMC} and a PID-based controller are designed. 
	The dynamic equations for the discussed systems are not affine in input, causing problems in control signal allocation to the actuators. To address this problem, we devised a solution discussed in subsection~\ref{subsec:affining} and completed in subsection~\ref{subsec:allocation}.
	The control scheme for the configurations Hedral, Tilt, and Half-Tilt, having four servo motors each, is a bit different from the configuration Tilt-Hedral which employs eight servo motors. However, we tried to comprehensively cover both in this section. 
	
%
	
	The state vector is described by
	\begin{equation}
		\begin{aligned}
			\bm{x} = \left[ \begin{array}{cc}
				\bm{\bar{x}} \\
				\dot{\bm{\bar{x}}}
				  \end{array} 
				  \right].
		\end{aligned}
	\end{equation}
	
	Control inputs for the configurations Hedral, Tilt, and Half-Tilt
	\begin{equation}
		\begin{aligned}
			\bm{u}
			=[\begin{array}{l l l l l l l l}
				\omega_1 & \omega_2 & \omega_3 & \omega_4 & \beta_1 & \beta_2 	& \beta_3 & \beta_4 \\
			\end{array}]^\intercal .
		\end{aligned}
	\end{equation}
	
	Control inputs for the configuration Tilt-Hedral
	\begin{equation}
		\begin{aligned}
			\bm{u}&=\begin{array}{l l l l l l l l l l l l}
				[\omega_1 & \omega_2 & \omega_3 & \omega_4 & \beta_1 & \beta_2 & \beta_3 & \beta_4 &\\ \gamma_1 & \gamma_2 & \gamma_3 & \gamma_4]^\intercal .
			\end{array} 
		\end{aligned}
	\end{equation}
	
	\subsection{Control Signal Transformation}
	\label{subsec:affining}
	To achieve an input-affine form of dynamic equations, we introduce a virtual control signal, represented by $\bm{v}$. Although this signal cannot be directly executed by the actuators, it enables us to address the input-non-affine nature of the dynamic system. In Subsection \ref{subsec:allocation} we will transform them to executable signals.
	
	 For the configurations Hedral, Tilt, and Half-Tilt
	\begin{equation}
		\label{affinev}
		\bm{v}=\left[\begin{array}{l}
			v_1 \\
			v_2 \\
			v_3 \\
			v_4 \\
			v_5 \\
			v_6 \\
			v_7 \\
			v_8
		\end{array}\right]=\left[\begin{array}{c}
			\omega_1^2 \sin \beta_1 \\
			\omega_2^2 \sin \beta_2 \\
			\omega_3^2 \sin \beta_3 \\
			\omega_4^2 \sin \beta_4 \\
			\omega_1^2 \cos \beta_1 \\
			\omega_2^2 \cos \beta_2 \\
			\omega_3^2 \cos \beta_3 \\
			\omega_4^2 \cos \beta_4
		\end{array}\right] .
	\end{equation}
	
\sethlcolor{yellow}

%

	For the configuration Tilt-Hedral
	\begin{equation}
		\label{affinev2}
		\bm{v}=\left[\begin{array}{l}
			v_1 \\
			v_2 \\
			v_3 \\
			v_4 \\
			v_5 \\
			v_6 \\
			v_7 \\
			v_8 \\
			v_9 \\
			v_{10} \\
			v_{11} \\
			v_{12} \\
		\end{array}\right]=\left[\begin{array}{c}
			\omega_1^2 \sin \beta_1 \\
			\omega_2^2 \sin \beta_2 \\
			\omega_3^2 \sin \beta_3 \\
			\omega_4^2 \sin \beta_4 \\
			\omega_1^2 \cos \beta_1 \sin \gamma_1 \\
			\omega_2^2 \cos \beta_2 \sin \gamma_2 \\
			\omega_3^2 \cos \beta_3 \sin \gamma_3 \\
			\omega_4^2 \cos \beta_4 \sin \gamma_4 \\
			\omega_1^2 \cos \beta_1 \cos \gamma_1 \\
			\omega_2^2 \cos \beta_2 \cos \gamma_2 \\
			\omega_3^2 \cos \beta_3 \cos \gamma_3 \\
			\omega_4^2 \cos \beta_4 \cos \gamma_4 \\
		\end{array}\right] .
	\end{equation}

%

By incorporating $\bm{v}$, the system dynamics can be expressed in the input-affine form as
\begin{equation}
	\label{affine}
	\ddot{{\bar{\bm{x}}}} =
	\bm{b}(\bm{x})+\bm{G}(\bm{\bar{x}})\bm{v}(\bm{u}) .
\end{equation}

where
\begin{equation}
	\bm{G}= \frac{\partial \ddot{\bm{\bar{x}}}}{\partial \bm{v}} \quad , \quad \bm{b}=\ddot{\bm{\bar{x}}}-\frac{\partial \ddot{\bm{\bar{x}}} }{\partial \bm{v}} \bm{v} .
\end{equation}


	
\subsection{Sliding Mode Controller}
A \ac{SMC} is utilized to achieve an effective control despite unmodeled dynamics, uncertainties, and disturbances.
		
	Let $\bm{\tilde{\bar{x}}}$ be the tracking error vector
	\begin{equation}
		\label{error vec}
		\bm{\tilde{\bar{x}}}=\bm{\bar{x}} - \bm{\bar{x}}_{d} ,
	\end{equation}
	where $\bm{\bar{x}}_d$ is the desired value of $\bm{\bar{x}}$ , and sliding surface defined as
	\begin{equation}
		\label{eq:sliding surfaces}
		\bm{s}=\bm{\dot{\tilde{{\bar{x}}}}} + \bm{\lambda} \pdot \bm{\tilde{\bar{x}}} ,
	\end{equation}
	where $\pdot$ denotes the element-wise production. $\bm{\lambda}$ is a vector with constant, strictly positive elements \cite{slotine}.
	
	The time derivative of $\bm{s}$ is
	\begin{equation}
		\label{sliding surfaces dot} 
		\bm{\dot{s}}=\bm{\ddot{\tilde{{\bar{x}}}}} + \bm{\lambda} \pdot \bm{\dot{\tilde{\bar{x}}}} .
	\end{equation}
	
	Satisfying the \textit{sliding condition} (\ref{sliding cond}) , as described by \citet{slotine}, leads to an asymptotically stable system due to its convergence to sliding surface $\bm{s}=0$.
	\begin{equation}
		\label{sliding cond}
		\frac{1}{2}\frac{d}{dt}s_i^2 \leq -\eta_i|s_i| , \quad i=1,2,...,6 ,
	\end{equation}
	where $s_i$ is the i-th element of $\bm{s}$ and $\eta_i$ is a strictly positive constant. To robustly satisfy this condition
	\begin{equation}
		\label{sdot-tanh}
		\bm{\dot{s}}=-\bm{k} \pdot \tanh{(\bm{\sigma} \pdot \bm{s})},
	\end{equation}
	where $\bm{\sigma}$ represents a vector of positive constant elements and $\bm{k}$ is a gain vector. In the absence of uncertainties, the components of this matrix should satisfy the condition $k_i>\eta_i$. However, these values should be increased if there's a need to compensate for uncertainties and disturbances.
	
	Combining (\ref{sliding surfaces dot}) and (\ref{sdot-tanh}):
	\begin{equation}
		\label{comb-tanh}
		\bm{\ddot{\tilde{{\bar{x}}}}} +  \bm{\lambda} \pdot \bm{\dot{\tilde{\bar{x}}}}=-\bm{k} \pdot \tanh{(\bm{\sigma} \pdot \bm{s})}.
	\end{equation}
	Substituting derivatives of (\ref{error vec}) into  (\ref{comb-tanh}):
	\begin{equation}
		\label{pre-main-control-eq}
		\bm{\ddot{\bar{x}}}=\bm{\ddot{\bar{x}}_d}+\bm{\lambda} \pdot  \bm{(\dot{\bar{x}}_d- \dot{\bar{x}})} -\bm{k} \pdot \tanh{(\bm{\sigma} \pdot \bm{s})} .
	\end{equation}
	Replacing (\ref{affine}) into (\ref{pre-main-control-eq})
	\begin{equation}
		\label{main-control-eq}
		\bm{Gv}=\bm{\ddot{\bar{x}}_d}+\bm{\lambda} \pdot \bm{(\dot{\bar{x}}_d- \dot{\bar{x}})} -\bm{b} -\bm{k} \pdot \tanh{(\bm{\sigma} \pdot \bm{s})}= \bm{w} .
	\end{equation}

	The robustness of the control system is enhanced by substituting the discontinuous function $\bm{k} \pdot sign(\bm{s})$ with $\bm{k} \pdot \tanh{(\bm{\sigma} \pdot \bm{s})}$. This modification leads to a small steady-state tracking error, which decreases as the elements of $\bm{\sigma}$ increase. Additionally, the control signals become smoother with larger values of $\bm{\sigma}$. This trade-off between tracking accuracy and signal smoothness can be adjusted to meet specific design objectives.
	
	
	\subsection{PID-Based Controller}
In this subsection, we introduce a novel PID-based controller. This design incorporates a gravity compensation component and combines a linear allocator and a non-linear allocator. The controller is developed with a focus on minimizing computational costs, making it suitable for deployment on low-cost microcontrollers. However, its capacity to minimize power consumption is lower compared to the \ac{SMC}.
 First, the dynamics is decoupled into six single-input single-output systems. This means that we set some virtual control inputs, denoted the set by the vector $\bm{q}$, each of which can not be executed independently with an available actuator. Each of these control inputs can directly be related to one of the control outputs of the system, enabling the direct use of six PID sub-controllers.
	
	\begin{figure*}
		\centering
		\includegraphics[width=0.98\linewidth]{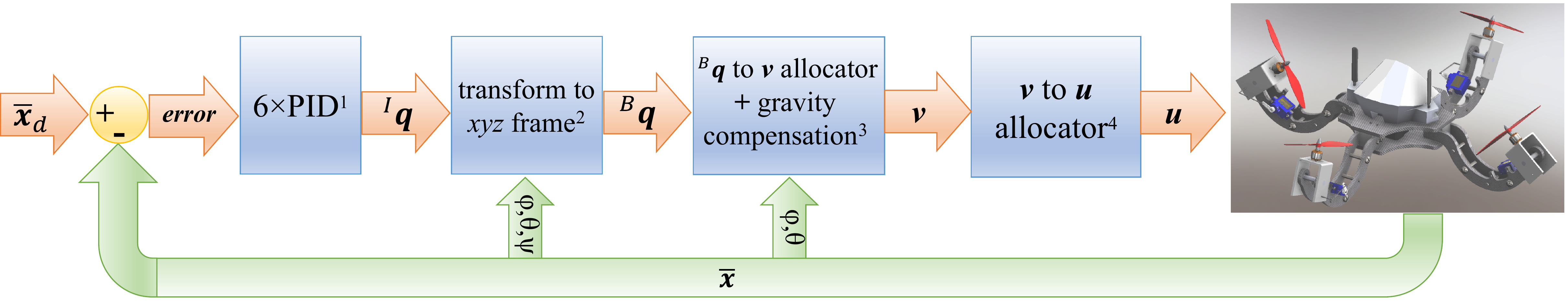}
		\caption{Concept of the proposed PID-based controller.\\ $^1$ Eq.~(\ref{pid}), $^2$ Eq.s~(\ref{trans q tot}),(\ref{trans q 1}),(\ref{trans q 2}), $^3$ Eq.s~(\ref{grav comp}),(\ref{w Gv}),(\ref{psudo2}), $^4$ Eq.~(\ref{ucomp}) or Eq.~(\ref{ucomp2})} 
		\label{fig:PID}
	\end{figure*}
	
	$\bm{q}$ is defined as
	\begin{equation}
		\leftindex^I{\bm{q}}=\left[\begin{array}{c}
			\leftindex^I{\bm{q}}_1 \\
			\leftindex^I{\bm{q}}_2 \end{array} \right] ,
	\end{equation}
	where
	\begin{equation}
		\leftindex^I{\bm{q}}_1=\left[\begin{array}{c}
			F_X \\
			F_Y \\
			F_Z \end{array} \right] \quad , \quad 	
		\leftindex^I{\bm{q}}_2=\left[\begin{array}{c}
			M_{x''} \\
			M_{y'} \\
			M_Z \end{array} \right] ,
	\end{equation}
	$F_X$, $F_Y$ and $F_Z$ are the sum of forces respectively in the inertia frame directions, $X$, $Y$, and $Z$. $M_{x''}$, $M_{y'}$ and $M_Z$ are the sum of moments respectively in $x''$, $y'$ and $Z$ directions.
	
	$\leftindex^I{\bm{q}}$ is obtained by the PID process
	\begin{equation}
		\label{pid}
		\leftindex^I{\bm{q}}=\bm{k_P} \pdot (\bar{\bm{x}_d}-\bar{\bm{x}})+\bm{k_D} \pdot (\dot{\bar{\bm{x}}}_d-\dot{\bar{\bm{x}}})+ \bm{k_I} \pdot \int (\bar{\bm{x}}_d-\bar{\bm{x}}) \,dt ,
	\end{equation}
	where $\bm{k_P}$, $\bm{k_D}$, and $\bm{k_I}$ are vectors with components of respectively, proportional, derivative, and integral gains.
	
Achieving a constant allocation matrix is desirable to avoid recalculating its elements and inversion in every iteration of real-time computations. To Accomplish this, $\bm{q}$ is expressed in $xyz$ frame as
	\begin{equation}
		\label{trans q tot}
		\leftindex^B{\bm{q}}=\left[\begin{array}{c}
			\leftindex^B{\bm{q}}_{1} \\
			\leftindex^B{\bm{q}}_{2} \end{array} \right] ,
	\end{equation}
	where $\leftindex^B{\bm{q}}_{1}$ and $\leftindex^B{\bm{q}}_{2}$ are obtained as
	\begin{equation}
		\label{trans q 1}
		\leftindex^B{\bm{q}}_{1}=\RO \leftindex^I{\bm{q}}_1 ,
	\end{equation}
	and
	\begin{equation}
		\label{trans q 2}
		\begin{aligned}
			\leftindex^B{\bm{q}}_{2} &=\Rx \left[ \begin{array}{c}
				M_{x''} \\
				0 \\
				0 \end{array} \right]
			+\Ry \Rx \left[ \begin{array}{c}
				0 \\
				M_{y'} \\
				0 \end{array} \right] \\
			& + \RO \left[ \begin{array}{c}
				0 \\
				0 \\
				M_Z \end{array} \right] .
		\end{aligned}
	\end{equation}
	
	$\leftindex^B{\bm{q}}$ denotes the sum of forces and moments, expressed in body frame, $xyz$. Thus
	\begin{equation}
		\leftindex^B{\bm{q}}=\bm{G}_b\bm{v}+\bm{c}_{gc} ,
	\end{equation}
	where we refer $\bm{c}_{gc}$ as the gravity compensation term
	\begin{equation}
		\label{grav comp}
		\bm{c}_{gc}= \left[ \begin{array}{c}
			mg \sin \theta \\
			-mg \cos \theta \sin \phi \\
			-mg \cos \theta \cos \phi \\
			0 \\
			0 \\
			0
		\end{array} \right] .
	\end{equation}
	$\bm{G}_b$ is unique for every configuration and derived by 

	\begin{equation}
		\bm{G}_b= \frac{ \partial \leftindex^B{\bm{h}} }
		{ \partial \bm{v}},
	\end{equation}
	where
	\begin{equation}
		\leftindex^B{\bm{h}}=\left[ \begin{array}{c}
			\leftindex^B{\bm{f}}\\
			\sum_{n=1}^4 \leftindex^B{\bm{r}}_{n / G} \times \leftindex^B{\bm{f}}_n+\sum_{n=1}^4 \leftindex^B{\bm{\tau}}_n 
		\end{array} \right],
	\end{equation}
	and the elements of $\leftindex^B{\bm{h}}$ are obtained by (\ref{f tot}), (\ref{moment tot}), and (\ref{tau tot}).

	For the Hedral configuration:	
	\begin{equation}
		\begin{aligned}
			&\bm{G}_b= \\
			&\left[ \begin{array}{cccccccc}
				k_f & k_f & 0 & 0 & 0 & 0 & 0 & 0 \\
				0 & 0 & -k_f & -k_f & 0 & 0 & 0 & 0 \\
				0 & 0 & 0 & 0 & k_f & k_f & k_f & k_f \\
				-k_m &k_m &ek_f &ek_f &0 &0 &-lk_f& lk_f \\
				ek_f & ek_f & k_m& -k_m &-lk_f& lk_f &0& 0 \\
				0& 0& 0& 0 &-k_m& k_m& -k_m& k_m
			\end{array} \right] .
		\end{aligned}
	\end{equation}
	
	For the Tilt configuration:
	\begin{equation}
		\begin{aligned}
			&\bm{G}_b= \\
			&\left[ \begin{array}{cccccccc}
				0&    0&   k_f&    k_f&     0&    0&     0&   0 \\
				-k_f&  -k_f&    0&     0&     0&    0&     0&    0 \\
				0&    0&    0&     0&    k_f&   k_f&    k_f&   k_f \\
				ek_f& ek_f&  -k_m&    k_m&     0&    0& -lk_f& lk_f \\
				k_m&  -k_m& ek_f&  ek_f& -lk_f& lk_f&     0&    0 \\
				-lk_f& lk_f& lk_f& -lk_f&   -k_m&   k_m&   -k_m&   k_m
			\end{array} \right] .
		\end{aligned}
	\end{equation}
	
	For the Tilt-Hedral configuration:
	\begin{equation}
		\begin{aligned}
			&\bm{G}_b= \\
			&\left[ \begin{array}{cccccccc}
				k_f&   k_f&    0&    0&     0&    0&   k_f&    k_f \\    
				0&    0&  -k_f&  -k_f&   -k_f&  -k_f&    0&     0  \\
				0&    0&    0&    0&     0&    0&    0&     0    \\
				-k_m&   k_m& ek_f& ek_f&  ek_f& ek_f&  -k_m&    k_m   \\
				ek_f& ek_f&   k_m&  -k_m&    k_m&  -k_m& ek_f&  ek_f \\
				0&    0&    0&    0& -lk_f& lk_f& lk_f& -lk_f  
			\end{array} \right. \\
			&\left.\begin{array}{cccc}
				0&    0&     0&    0 \\
				0&    0&     0&    0\\
				k_f&   k_f&    k_f&   k_f\\
				0&    0& -lk_f& lk_f\\
				-lk_f& lk_f&     0&    0\\
				-k_m&   k_m&   -km&   km \end{array}   \right] .
		\end{aligned}
	\end{equation}
	
	For the Half-Tilt configuration, $\bm{G}_b$ is a rank-defficient matrix, so it is not mentioned here. Finally
	\begin{equation}
		\label{w Gv}
		\bm{G}_b \bm{v}=\leftindex^B{\bm{q}}-\bm{c}_{gc}=\bm{w}_b .
	\end{equation}
	Figure~\ref{fig:PID} visually demonstrates the concept of this controller.

	\subsection{Control Allocation}
	\label{subsec:allocation}
	The equation $\bm{G}_{n \times m} \bm{v}_{m \times 1}=\bm{w}_{n \times 1}$ represents an under-determined system since $m>n$. If $\bm{G}$ is full rank, this equation has infinitely many solutions. The \textit{minimum-normed solution} provided by (\ref{psudo1}) minimizes the norm of $\bm{v}$ (\ref{normv}). This strategy contributes to reducing power consumption and extended battery life by minimizing the inputs to BLDS motors.
	
\begin{equation}
	\label{normv}
	\|\bm{v}\|=(v_1^2+v_2^2+...+v_m^2)^{1/2}=(\omega_1^4+\omega_2^4+\omega_3^4+\omega_4^4)^{1/2} .
\end{equation}	

$\bm{G}$ and $\bm{G}_b$ are full rank for all the configurations, except the Half-Tilt configuration. As a result, an exact solution for $\bm{v}$ is not attainable, meaning that some of actuators are not independent in Half-Tilt configuration and this system remains under-actuated. 
A closer observation of its dynamics reveals that all thrust vector rotations occur solely around the $x$-axis. As a result, one pair of servo motors does not contribute an additional level of actuation, since this configuration cannot independently control its pitch angle relative to translation along the $x$-axis.

 Thus, this configuration does not meet our desired goal of achieving complete independent control over all six \ac{DoF} and it will not be discussed in the continue.

	For the sliding mode controller:
	\begin{equation}
		\label{psudo1}
		\bm{v}=\bm{G}^T(\bm{G} \bm{G}^T )^{-1} \bm{w} .
	\end{equation}
%
	For the PID-based controller:
	\begin{equation}
		\label{psudo2}
		\bm{v}=\bm{G}_b^T(\bm{G}_b \bm{G}_b^T )^{-1} \bm{w}_b .
	\end{equation}
	Note that in the PID-based controller, the allocation matrix $\bm{G}_b$ is constant and degrades some of the capabilities for minimizing power usage, while \ac{SMC} computes $\bm{G}$ and a minimum normed solution of $\bm{v}$ in every loop of calculations.
	
Eventually, the main control signals are calculated according to (\ref{ucomp}) and (\ref{ucomp2}) as the elements of vector $\bm{u}$.
	
	For the configurations Hedral, Tilt, and Half-Tilt:
		\begin{equation}
			\label{ucomp}
			\begin{aligned}
				& \beta_i=\tan^{-1}(\frac{v_i}{v_{i+4}}), \\		
				& \omega_i=(v_i^2+v_{i+4}^2)^{1/4} \quad , i=1,2,3,4.			
			\end{aligned}
		\end{equation}

		For the configuration Tilt-Hedral:
		\begin{equation}
			\label{ucomp2}
			\begin{aligned}
				&\gamma_i= \tan^{-1}(\frac{v_{i+4}}{v_{i+8}}), \\
				&\beta_i= \tan^{-1}(\frac{v_i}{v_{i+8}}) \cos\gamma_i, \\
				& \omega_i=(v_i^2+v_{i+4}^2+v_{i+8}^2)^{1/4} \quad , i=1,2,3,4.\\		
				&
			\end{aligned}
		\end{equation}
		
		%
		
		\section{Simulation and Results}
		\label{sec:simulation}
		
To facilitate the numerical computation of the system's differential equations, they are represented in the following state-space form as
\begin{equation}
	\bm{\dot{x}}=\left[\begin{array}{c}
		\dot{{\bar{\bm{x}}}} \\
		\ddot{{\bar{\bm{x}}}}
	\end{array}\right]. 
\end{equation}
		
The following four distinct maneuvers are analyzed to show the exceptional capabilities of the presented configurations, which surpass the abilities of a standard multi-rotor.

\begin{figure}
	\centering
	\includegraphics[width=0.99\linewidth]{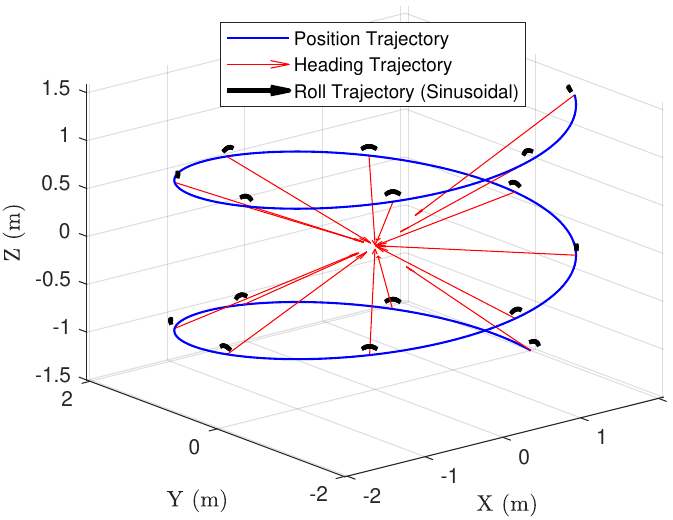}
	\caption{Trajectory of Maneuver 4.} 
	\label{fig:helix}
\end{figure}

\begin{itemize}
	\item  
	\textbf{Maneuver 1}: Sinusoidal translation without rotation of the body,
	\begin{equation}
		\begin{aligned}
			& X_d=1 \sin(0.4t), \quad Y_d=1 \sin(0.4t), \quad Z_d=2,\\
			& \phi_d=0, \quad \theta_d=0, \quad \psi_d=0 .
		\end{aligned}
	\end{equation}
	
	\item 
	\textbf{Maneuver 2}: Hover with a desired attitude,
	\begin{equation}
		\begin{aligned}
			& X_d=0, \quad Y_d=0, \quad Z_d=2,\\
			& \phi_d=20^\circ, \quad \theta_d=25^\circ, \quad \psi_d=90^\circ .
		\end{aligned}
	\end{equation}
	
	\item 
	 \textbf{Maneuver 3}: Simultaneous sinusoidal translation and rotation,
	\begin{equation}
		\begin{aligned}
			 &X_d=1 \sin(0.4t)&, \quad Y_d=1 \sin(0.4t),\\  \quad & Z_d=1\sin(0.4t)&,\quad
			 \phi_d=20^\circ \sin(0.3t), \\  &\theta_d=20^\circ \sin(0.3t)&, \quad \psi_d=20^\circ \sin(0.3t) .
		\end{aligned}
	\end{equation}
	
	\item 
	\textbf{Maneuver 4}: The drone performs a helical position trajectory, maintaining its heading focused on a central point in the helix while following a sinusoidal roll pattern with a $30^\circ$
	amplitude. The helix has a radius of 2 meters and a height of 3 meters, and the drone forced to complete the maneuver within 30 seconds. Figure~\ref{fig:helix} illustrates this trajectory.
	
\end{itemize}

The simulations are performed under two conditions: without and with uncertainties and disturbances. To ensure a concise presentation, we do not include the results for every maneuver across all configuration and controller combinations, instead focusing on key outcomes. Tables \ref{tab:ideal figures} and \ref{tab:real figures} provide references to figures showcasing the simulation results under these two conditions.
		
%

			{\renewcommand{\arraystretch}{1.5}
	\begin{table}[h]
		\centering
		\fontsize{9pt}{9pt}\selectfont
		\caption{References to figures illustrating simulation results without uncertainties and disturbances. Cells without specific maneuver indications correspond to simulations for maneuver~3.}
		\label{tab:ideal figures}
		\begin{tabular}{|l|c|c|c|}
			\hline
			Controller& Hedral & Tilt & Tilt-Hedral \\
			\hline
			SMC & \makecell{maneuver1:Fig. \ref{fig:hedral-1} \\ maneuver2: Fig. \ref{fig:hedral-2} \\ maneuver3: Fig. \ref{fig:hedral-3} \\  maneuver4: Fig. \ref{fig:hedral-ideal-4} } & Fig. \ref{fig:tilt-1} & Fig. \ref{fig:tilthedral-1} \\
			\hline
			
			PID-based & Fig. \ref{fig:hedral-PID-1} & Fig. \ref{fig:tilt-PID-1} & Fig. \ref{fig:tilthedral-PID-1} \\
			\hline
		\end{tabular}
	\end{table}
}

			{\renewcommand{\arraystretch}{1.5}
			\begin{table}[h]
				\centering
				\fontsize{9pt}{9pt}\selectfont
				\caption{References to figures illustrating simulation results in the presence of uncertainties and disturbances. Cells without specific maneuver indications correspond to simulations for maneuver~3.}
				\label{tab:real figures}
				\begin{tabular}{|l|c|c|c|}
					\hline
					Controller& Hedral & Tilt & Tilt-Hedral \\
					\hline
					SMC & \makecell{maneuver1:Fig. \ref{fig:hedral-real-1} \\ maneuver2: Fig. \ref{fig:hedral-real-2} \\ maneuver3: Fig. \ref{fig:hedral-real-3} \\  maneuver4: Fig. \ref{fig:hedral-real-4} } & Fig. \ref{fig:tilt-real-1} & Fig. \ref{fig:tilthedral-real-1} \\
					\hline
					
					PID-based & Fig. \ref{fig:hedral-real-PID-3} & Fig. \ref{fig:tilt-real-PID-3} & Fig. \ref{fig:tilthedral-real-PID-3} \\
					\hline
				\end{tabular}
			\end{table}
		}
		
		\subsection{Simulation Without Uncertainties and Disturbances}
		In this scenario, we apply the controller to the dynamic model exactly coincides with the model incorporated in the controller, meaning there are no uncertainties and disturbances. The values of parameters for the simulations are shown in Table \ref{simparams}. 		
		\begin{table}[h]
			\caption{Parameter values for simulations.}
			\centering
			\fontsize{9pt}{9pt}\selectfont
			\label{simparams}
			\begin{tabular}{|c|c|c|}
				\hline
				Parameter & Value & Unit \\
				\hline
				$g$ & $9.81$ & $m/s^2$ \\
				$m$ & $1$ & $kg$ \\
				$l$ & $0.3$ & $m$ \\
				$e$ & $0.05$ & $m$ \\
				$I_{xx}$ & $10 \times 10^{-3}$ & $kg.m^2$ \\
				$I_{yy}$ & $10\times10^{-3}$ & $kg.m^2$ \\
				$I_{zz}$ & $17\times10^{-3}$ & $kg.m^2$ \\
				$k_f$ & $6\times10^{-6}$ & $N.s^2$ \\
				$k_m$ & $4\times10^{-7}$ & $N.m.s^2$ \\ 
				\hline
			\end{tabular}
		\end{table}

\begin{figure}
	\centering
		\begin{subfigure}{0.95\linewidth}
			\includegraphics[width=7.5cm]{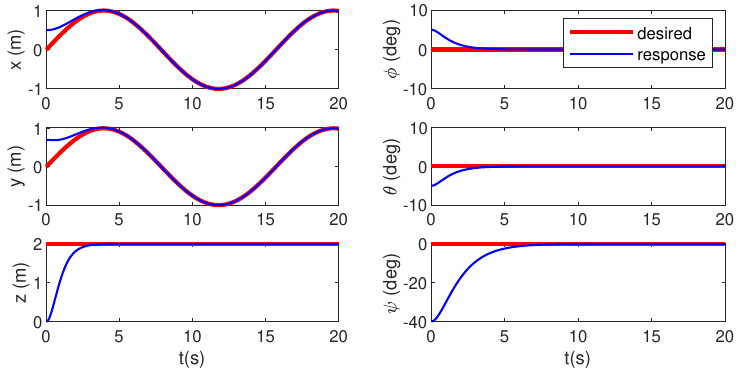}
			\caption{Position and Orientation Tracking} 
		\end{subfigure}
		\vspace{2ex}
		\begin{subfigure}{0.95\linewidth}
			\includegraphics[width=7.5cm]{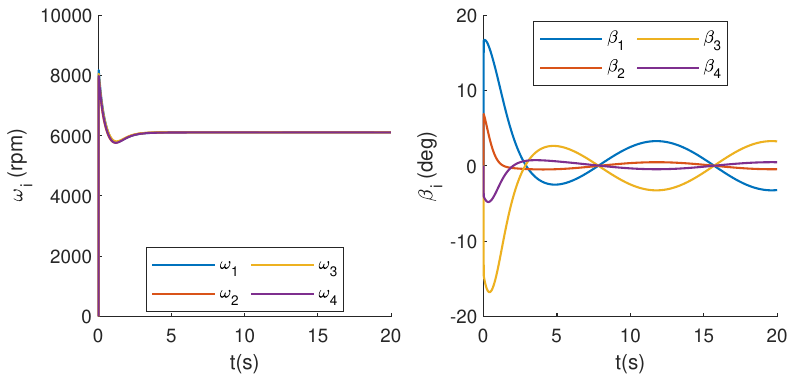}
			\caption{Control Inputs} 
		\end{subfigure}
		
		\caption{
			Simulation of the maneuver 1 for Hedral configuration using SMC without uncertainties and disturbances. 
		} 
		
		\label{fig:hedral-1}
\end{figure}

\begin{figure}
		\begin{subfigure}{0.95\linewidth}
			\includegraphics[width=7.5cm]{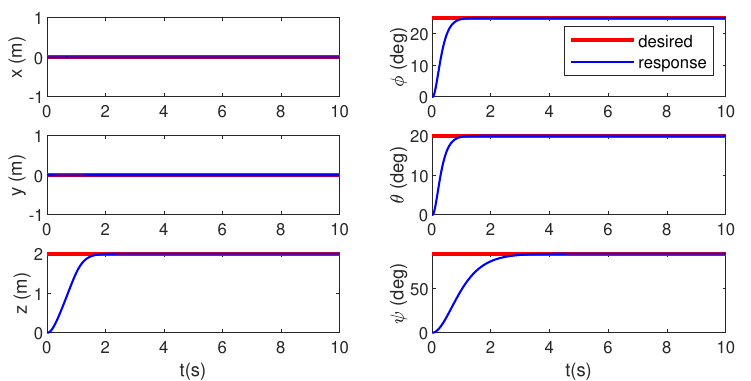}
			\caption{Position and Orientation Tracking} 
		\end{subfigure}
		\vspace{2ex}
		\begin{subfigure}{0.95\linewidth}
			\includegraphics[width=7.5cm]{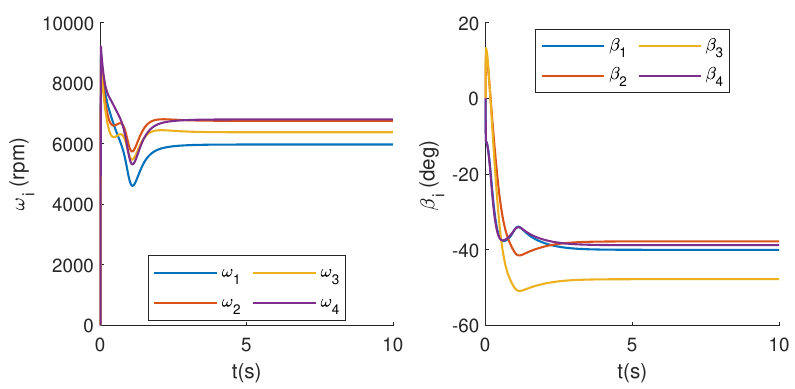}
			\caption{Control Inputs} 
		\end{subfigure}
		
		\caption{Simulation of the maneuver 2 for Hedral configuration using SMC without uncertainties and disturbances.} 
			\label{fig:hedral-2}
\end{figure}


\begin{figure}
	\centering
	\begin{subfigure}{\linewidth}
		\includegraphics[width=\linewidth]{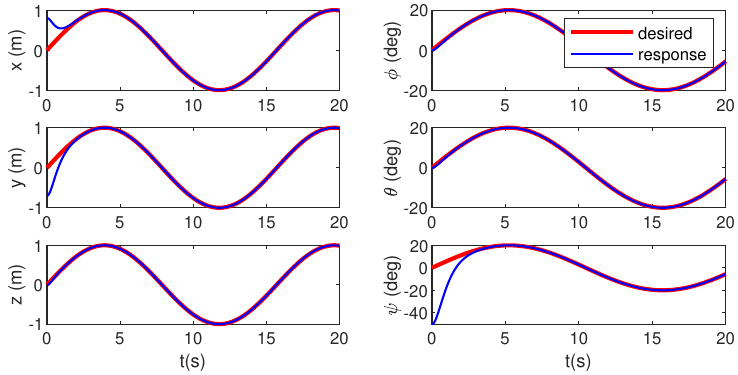}
		\caption{Position and Orientation Tracking} 
	\end{subfigure}
	\vspace{2ex}
	\begin{subfigure}{\linewidth}
		\includegraphics[width=\linewidth]{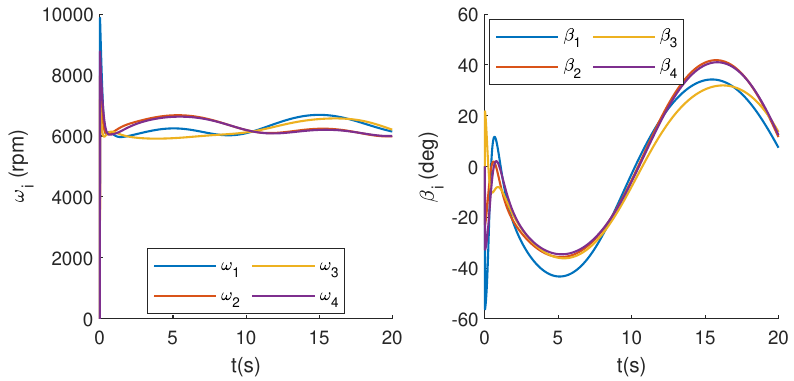}
		\caption{Control Inputs} 
	\end{subfigure}
	
	\caption{Simulation of the maneuver 3 for Hedral configuration using SMC without uncertainties and disturbances.} 
	\label{fig:hedral-3}
\end{figure}

\begin{figure}
		\begin{subfigure}{0.99\linewidth}
			\includegraphics[width=\linewidth]{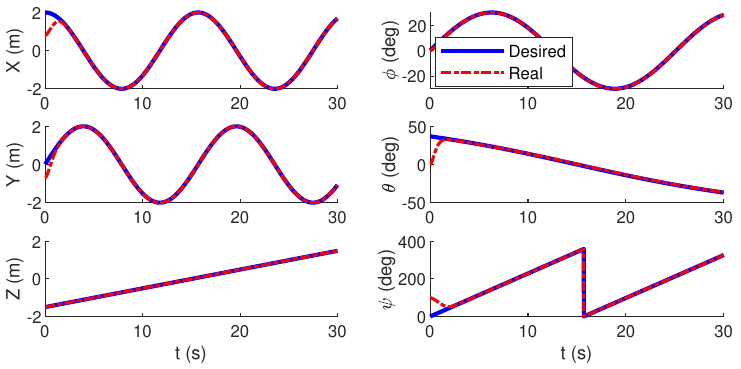}
			\caption{Position and Orientation Tracking} 
		\end{subfigure}
		\vspace{2ex}
		\begin{subfigure}{\linewidth}
			\includegraphics[width=0.99\linewidth]{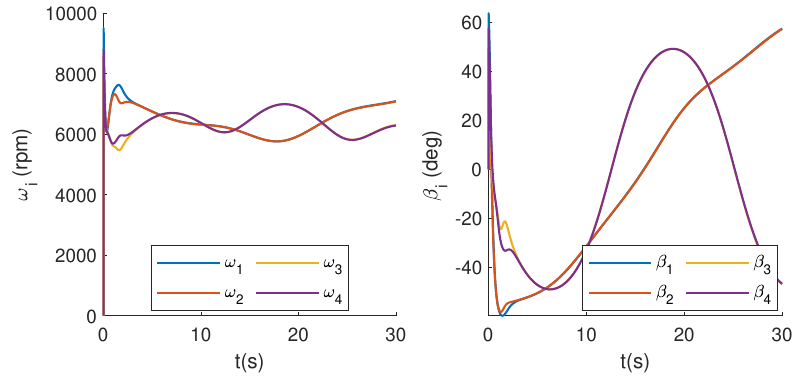}
			\caption{Control Inputs} 
		\end{subfigure}
		
		\caption{Simulation of the maneuver 4 for Hedral configuration using SMC without uncertainties and disturbances.}
		\label{fig:hedral-ideal-4}
\end{figure}

\begin{figure}
	\begin{subfigure}{0.99\linewidth}
		\includegraphics[width=\linewidth]{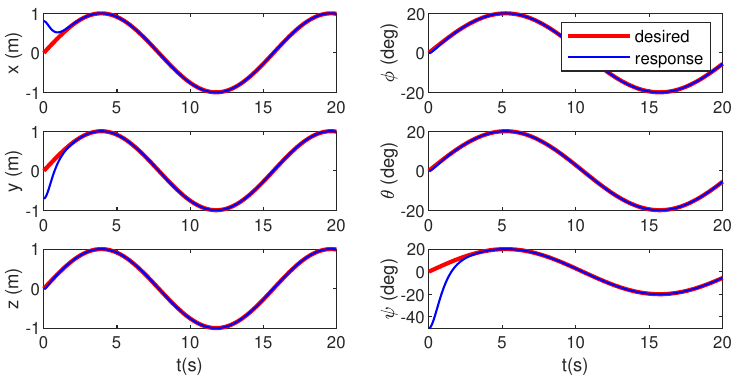}
		\caption{Position and Orientation Tracking} 
	\end{subfigure}
	\vspace{2ex}
	\begin{subfigure}{0.99\linewidth}
		\includegraphics[width=\linewidth]{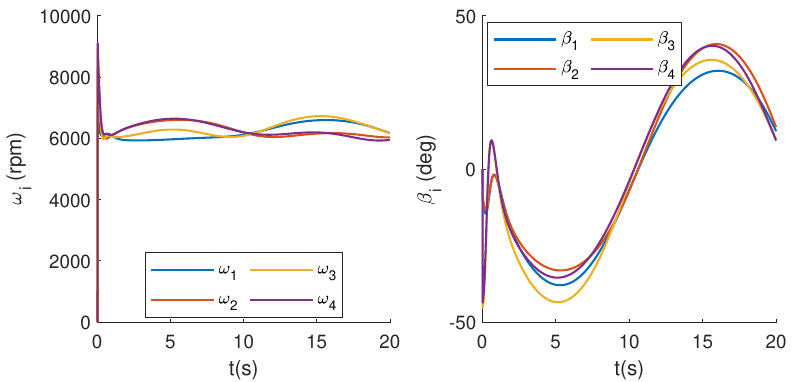}
		\caption{Control Inputs} 
	\end{subfigure}
	
	\caption{Simulation of the maneuver 3 for Tilt configuration using SMC without uncertainties and disturbances.} 
	\label{fig:tilt-1}
\end{figure}

\begin{figure}
	\begin{subfigure}{0.95\linewidth}
		\includegraphics[width=\linewidth]{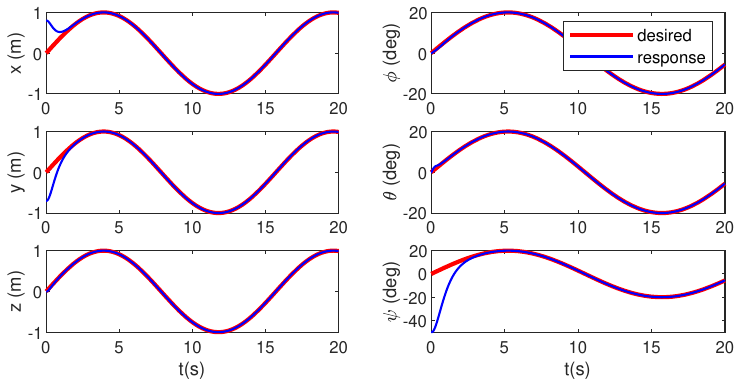}
		\caption{Position and Orientation Tracking} 
	\end{subfigure}
	\vspace{2ex}
	\begin{subfigure}{0.95\linewidth}
		\includegraphics[width=\linewidth]{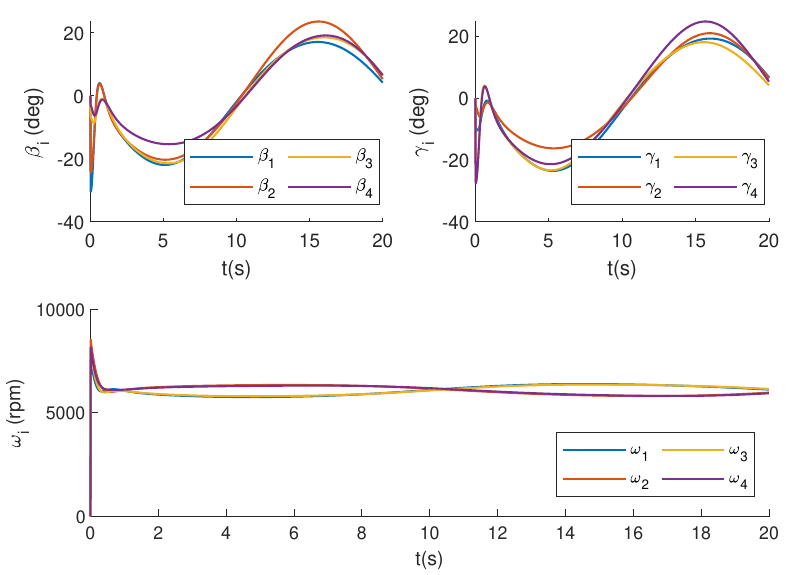}
		\caption{Control Inputs} 
	\end{subfigure}
	
	\caption{Simulation of the maneuver 3 for Tilt-Hedral configuration using SMC without uncertainties and disturbances.} 
	\label{fig:tilthedral-1}
\end{figure}

\begin{figure}
	\begin{subfigure}{0.95\linewidth}
		\includegraphics[width=\linewidth]{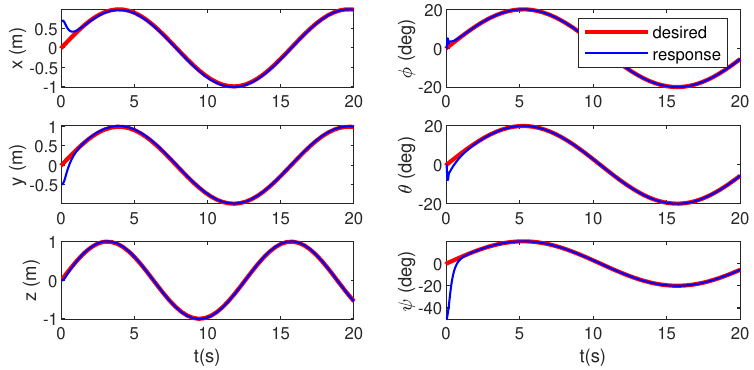}
		\caption{Position and Orientation Tracking} 
	\end{subfigure}
	\vspace{2ex}
	\begin{subfigure}{0.95\linewidth}
		\includegraphics[width=\linewidth]{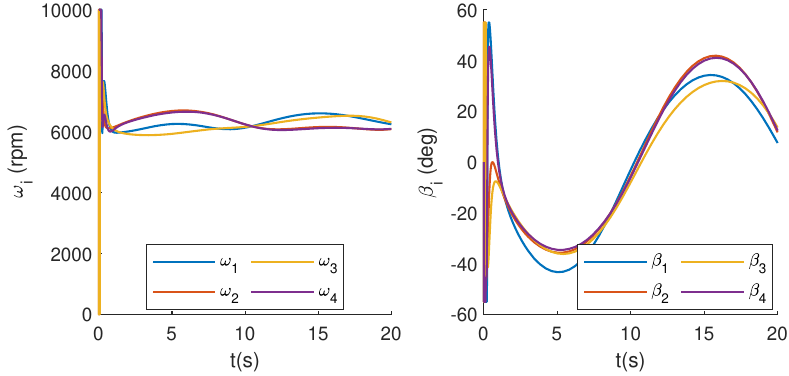}
		\caption{Control Inputs} 
	\end{subfigure}
	
	\caption{Simulation of the maneuver 3 for Hedral configuration using PID-based controller without uncertainties and disturbances.} 
	\label{fig:hedral-PID-1}
\end{figure}

\begin{figure}
	\begin{subfigure}{0.95\linewidth}
		\includegraphics[width=\linewidth]{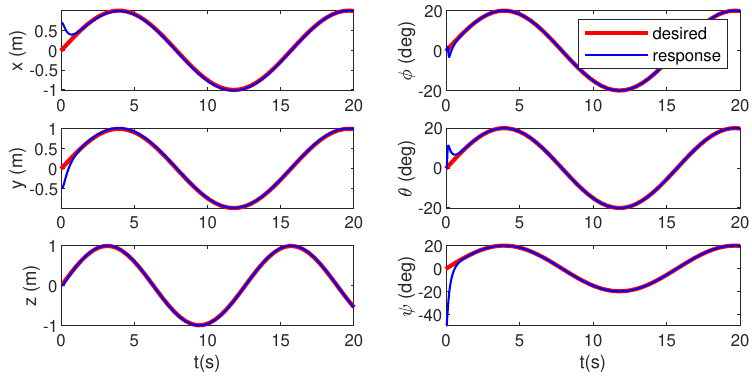}
		\caption{Position and Orientation Tracking} 
	\end{subfigure}
	\vspace{2ex}
	\begin{subfigure}{0.95\linewidth}
		\includegraphics[width=\linewidth]{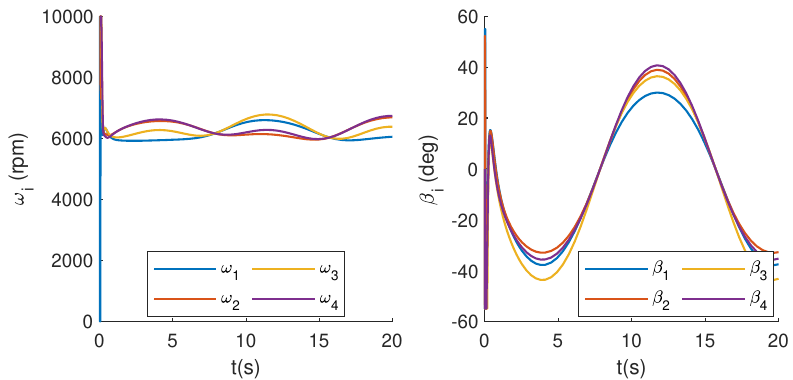}
		\caption{Control Inputs} 
	\end{subfigure}
	
	\caption{Simulation of the maneuver 3 for Tilt configuration using PID-based controller without uncertainties and disturbances.} 
	\label{fig:tilt-PID-1}
\end{figure}

\begin{figure}
	\begin{subfigure}{0.95\linewidth}
		\includegraphics[width=\linewidth]{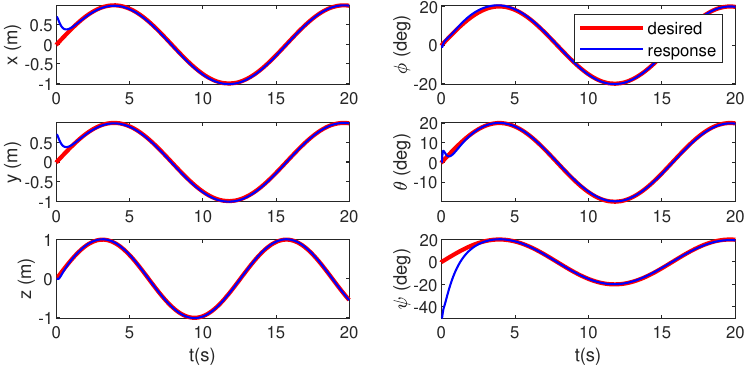}
		\caption{Position and Orientation Tracking} 
	\end{subfigure}
	\vspace{2ex}
	\begin{subfigure}{0.95\linewidth}
		\includegraphics[width=\linewidth]{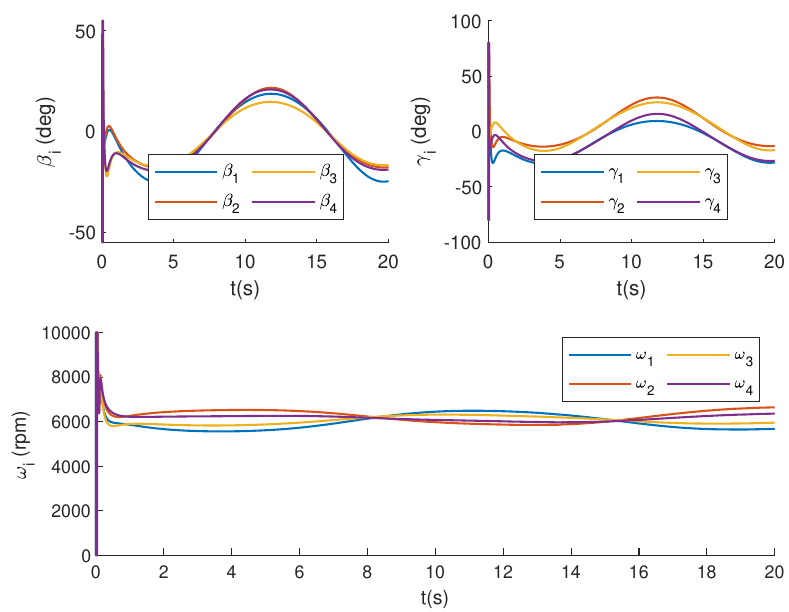}
		\caption{Control Inputs} 
	\end{subfigure}
	
	\caption{Simulation of the maneuver 3 for Tilt-Hedral configuration using PID-based controller without uncertainties and disturbances.} 
	\label{fig:tilthedral-PID-1}
\end{figure}		
		
		\subsection{Robustness Study}
		\label{subsec:realsim}
		We introduce the following uncertainties and disturbances to challenge the robustness of the designed control systems.

\textbf{a)} A first-order delay of 100 $\mu s$ for BLDC motors and a delay of 0.4 $ deg/60s$ seconds for servo motors.

\textbf{b)} Actuation errors of 1-5\% for servo motors and 4-10\% for BLDC motors. 

\textbf{c)} Uncertainties in physical parameters: mass increased by 5\% in the realistic model, inertia moments increased by 20\%, aerodynamic coefficients of propellers reduced by 10\%, and mass center eccentricity increased by 20\%.

\textbf{d)} Random disturbance forces (e.g., wind) with a resultant magnitude ranging from 5–10\% of the system's weight are applied in the X and Y directions at a frequency of 100 Hz. 
 (\ref{dyn disturb}) represents the implementation of this disturbance in the system's dynamics.
\begin{equation}
	\label{dyn disturb}
	\ddot{{\bar{\bm{x}}}} =\left[ \begin{array}{c}
		\ddot{\bm{p}} \\
		\ddot{\bm{o}}
	\end{array} \right]+\bm{a}_d,
\end{equation}
where 
\begin{equation}
	\label{cd}
	\bm{a}_d=\left[ \begin{array}{cccccc}
		\xi_X g &\xi_Y g & 0 & 0 & 0 & 0 \end{array} \right]^\intercal ,	
\end{equation}
and $\xi_X$ and $\xi_Y$ are randomly varying parameters in the range of 3.53-7.07\% , yielding a resultant of 5-10\%.

 As the results indicate, the controllers manage to overcome the aforementioned uncertainties and track the desired trajectories. It's important to notice that increasing the gains of the \ac{SMC} minimizes the tracking error, but leads to high chattering and overshoot in control inputs which is not desirable in real-world scenarios.

		\begin{figure}
			\begin{subfigure}{0.95\linewidth}
				\includegraphics[width=7.5cm]{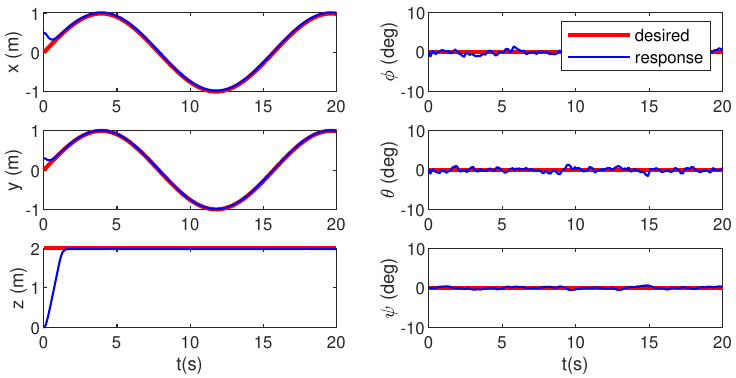}
				\caption{Position and Orientation Tracking} 
			\end{subfigure}
			\vspace{2ex}
			\begin{subfigure}{0.95\linewidth}
				\includegraphics[width=7.5cm]{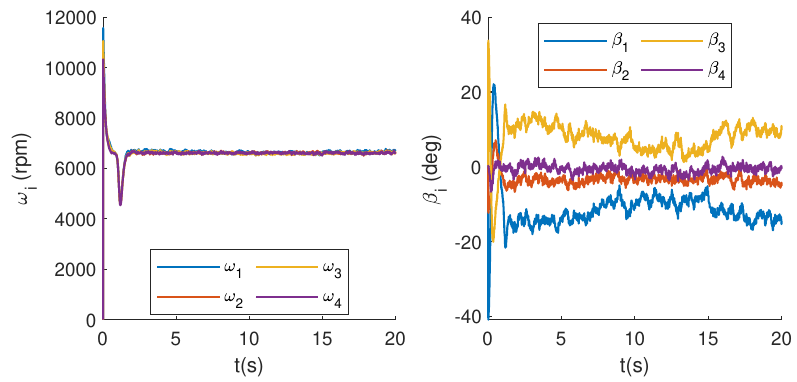}
				\caption{Control Inputs} 
			\end{subfigure}
			
			\caption{Simulation in presence of uncertainties and disturbances of the maneuver 1 for Hedral configuration using SMC.} 
			\label{fig:hedral-real-1}
		\end{figure}
		
		\begin{figure}
			\begin{subfigure}{0.95\linewidth}
				\includegraphics[width=7.5cm]{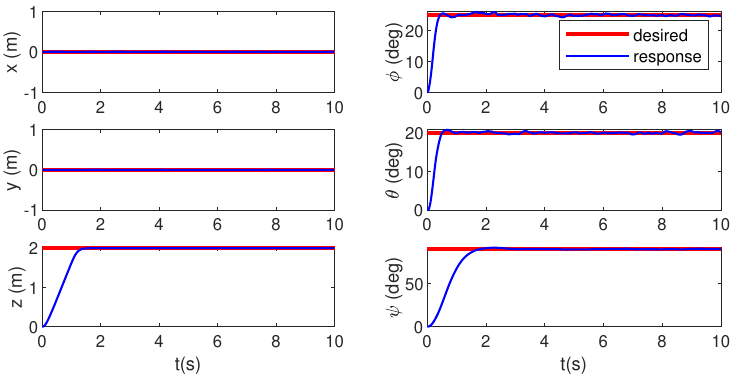}
				\caption{Position and Orientation Tracking} 
			\end{subfigure}
			\vspace{2ex}
			\begin{subfigure}{0.95\linewidth}
				\includegraphics[width=7.5cm]{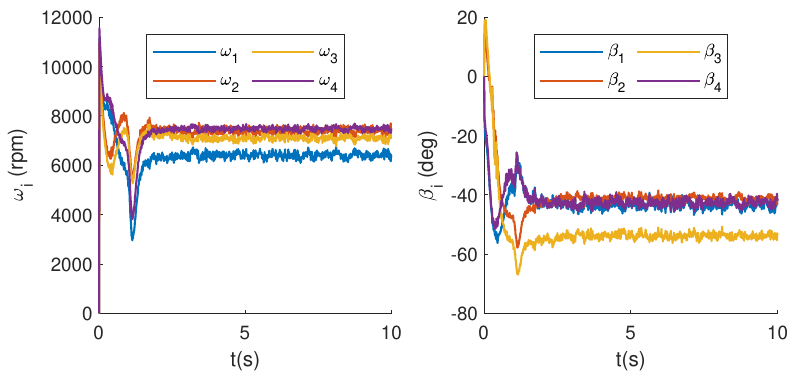}
				\caption{Control Inputs} 
			\end{subfigure}
			
			\caption{Simulation in presence of uncertainties and disturbances of the maneuver 2 for Hedral configuration using SMC.} 
			\label{fig:hedral-real-2}
		\end{figure}
		
		\begin{figure}
			\begin{subfigure}{0.95\linewidth}
				\includegraphics[width=7.5cm]{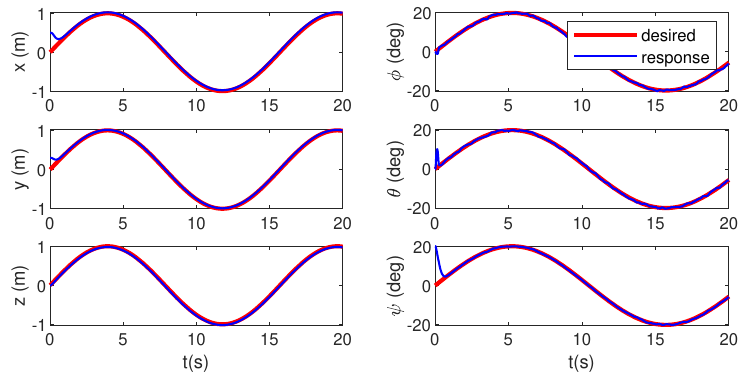}
				\caption{Position and Orientation Tracking} 
			\end{subfigure}
			\vspace{2ex}
			\begin{subfigure}{0.95\linewidth}
				\includegraphics[width=7.5cm]{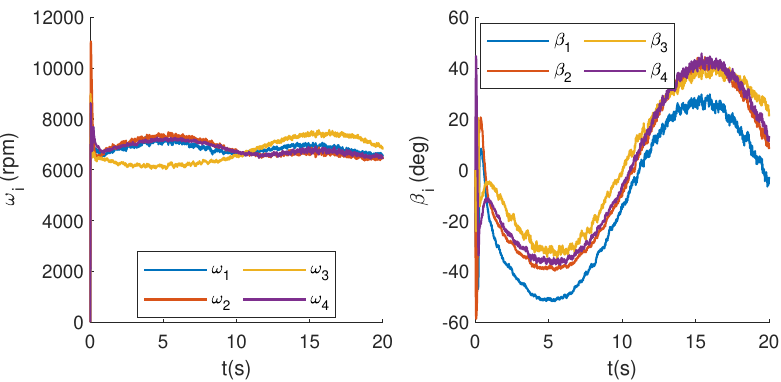}
				\caption{Control Inputs} 
			\end{subfigure}
			
			\caption{Simulation in presence of uncertainties and disturbances of the maneuver 3 for Hedral configuration using SMC.} 
			\label{fig:hedral-real-3}
		\end{figure}
		
		\begin{figure}
			\begin{subfigure}{0.95\linewidth}
				\includegraphics[width=7.5cm]{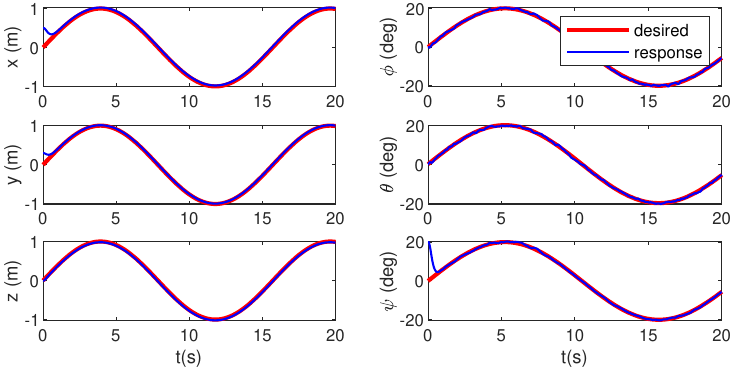}
				\caption{Position and Orientation Tracking} 
			\end{subfigure}
			\vspace{2ex}
			\begin{subfigure}{0.95\linewidth}
				\includegraphics[width=7.5cm]{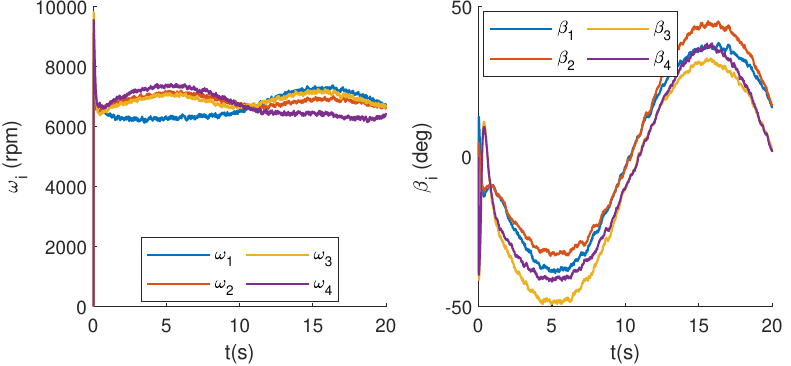}
				\caption{Control Inputs} 
			\end{subfigure}
			
			\caption{Simulation in presence of uncertainties and disturbances of the maneuver 3 for Tilt configuration using SMC.} 
			\label{fig:tilt-real-1}
		\end{figure}
		
		\begin{figure}
			\begin{subfigure}{0.95\linewidth}
				\includegraphics[width=7.5cm]{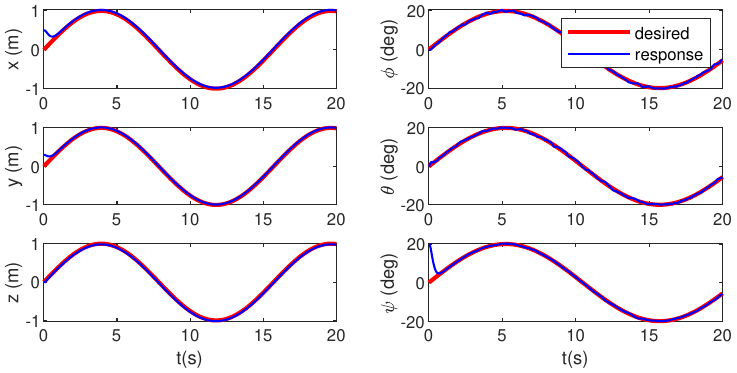}
				\caption{Position and Orientation Tracking} 
			\end{subfigure}
			\vspace{2ex}
			\begin{subfigure}{0.95\linewidth}
				\includegraphics[width=7.5cm]{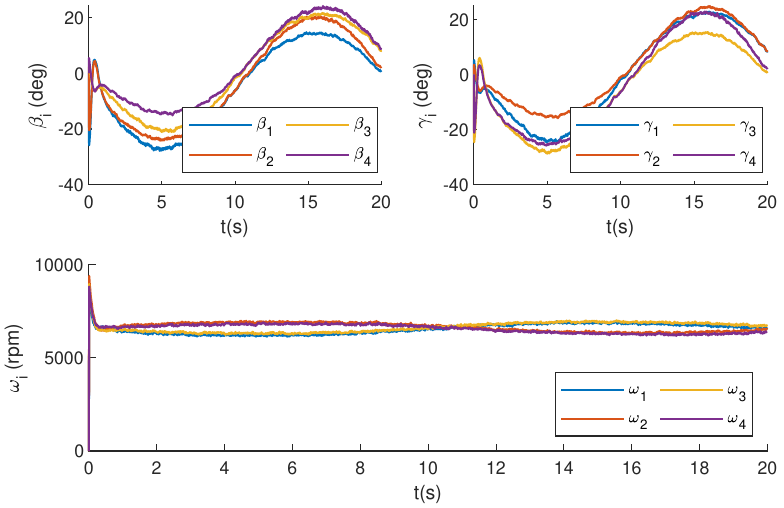}
				\caption{Control Inputs} 
			\end{subfigure}
			
			\caption{Simulation in presence of uncertainties and disturbances of the maneuver 3 for Tilt-Hedral configuration using SMC.} 
			\label{fig:tilthedral-real-1}
		\end{figure}
		
		\begin{figure}
			\begin{subfigure}{0.95\linewidth}
				\includegraphics[width=7.5cm]{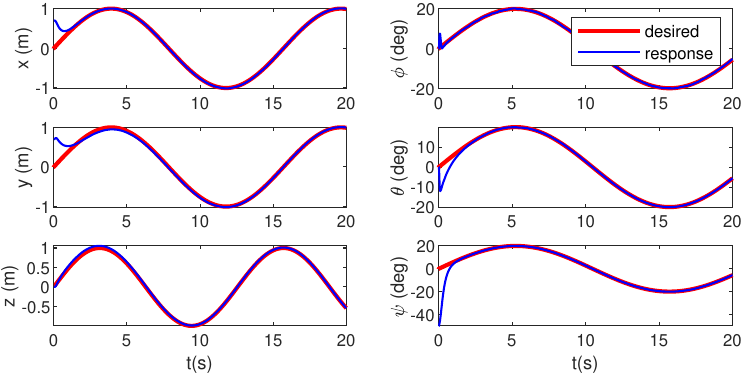}
				\caption{Position and Orientation Tracking} 
			\end{subfigure}
			\vspace{2ex}
			\begin{subfigure}{0.95\linewidth}
				\includegraphics[width=7.5cm]{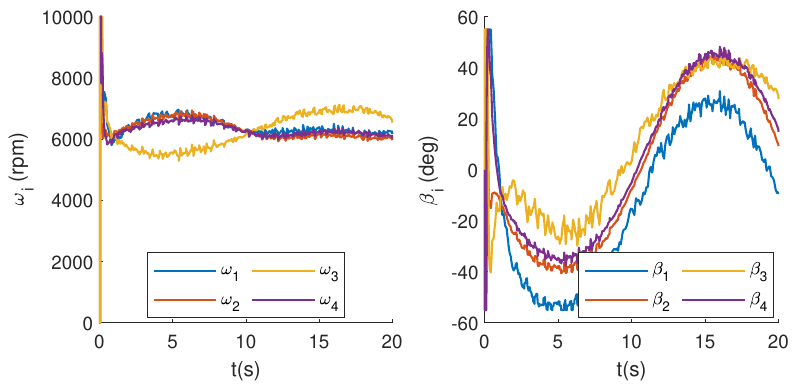}
				\caption{Control Inputs} 
			\end{subfigure}
			
			\caption{Simulation in presence of uncertainties and disturbances of the maneuver 3 for Hedral configuration using PID-based controller.} 
			\label{fig:hedral-real-PID-3}
		\end{figure}
		
		\begin{figure}
			\begin{subfigure}{0.95\linewidth}
				\includegraphics[width=7.3cm]{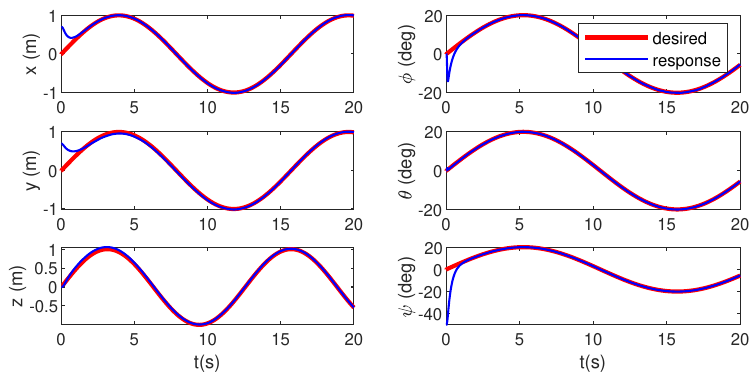}
				\caption{Position and Orientation Tracking} 
			\end{subfigure}
			\vspace{2ex}
			\begin{subfigure}{0.95\linewidth}
				\includegraphics[width=7.3cm]{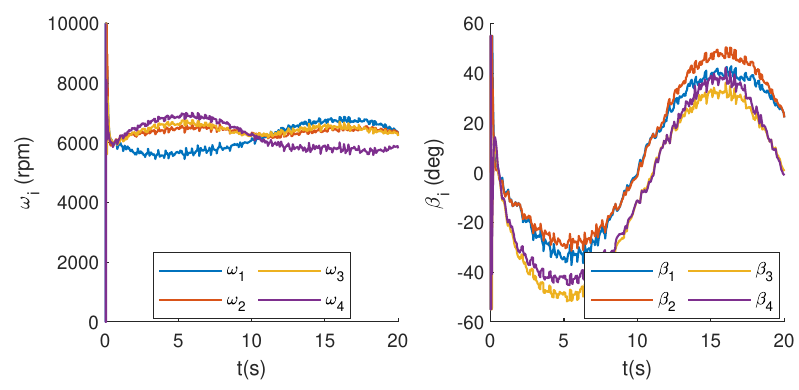}
				\caption{Control Inputs} 
			\end{subfigure}
			
			\caption{Simulation in presence of uncertainties and disturbances of the maneuver 3 for Tilt configuration using PID-based controller.} 
			\label{fig:tilt-real-PID-3}
		\end{figure}
		
		\begin{figure}
			\begin{subfigure}{0.9\linewidth}
				\includegraphics[width=\linewidth]{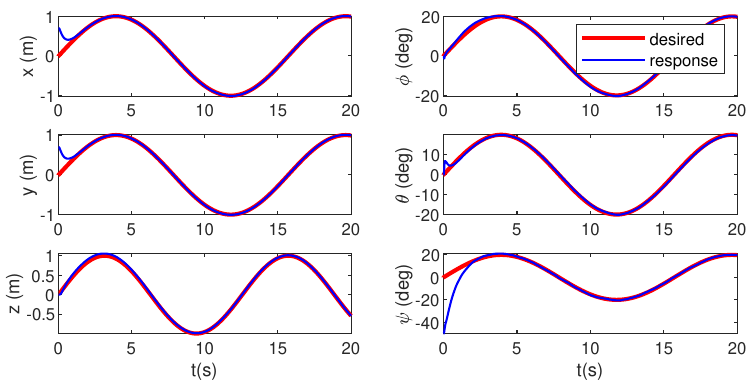}
				\caption{Position and Orientation Tracking} 
			\end{subfigure}
			\vspace{2ex}
			\begin{subfigure}{\linewidth}
				\includegraphics[width=0.9\linewidth]{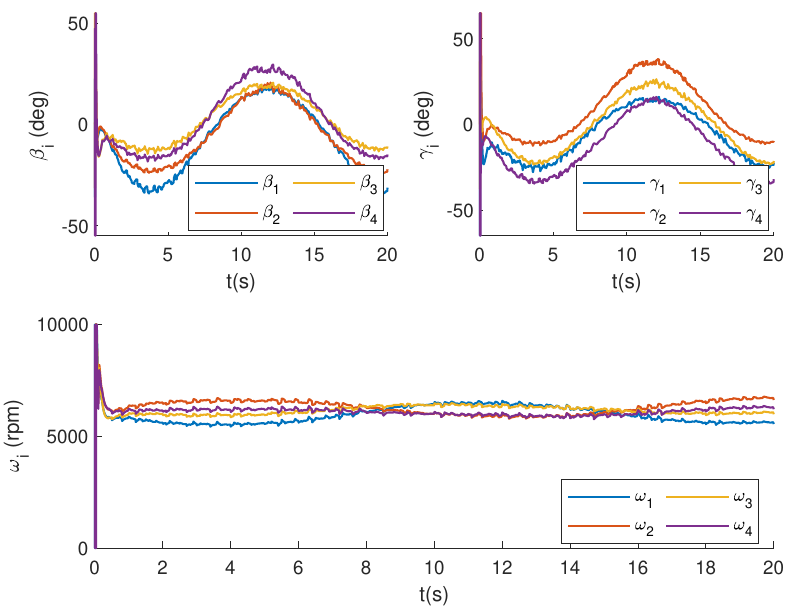}
				\caption{Control Inputs} 
			\end{subfigure}
			
			\caption{Simulation in presence of uncertainties and disturbances of the maneuver 3 for Tilt-Hedral configuration using PID-based controller.} 
			\label{fig:tilthedral-real-PID-3}
		\end{figure}

\begin{figure}
			\begin{subfigure}{0.95\linewidth}
				\includegraphics[width=\linewidth]{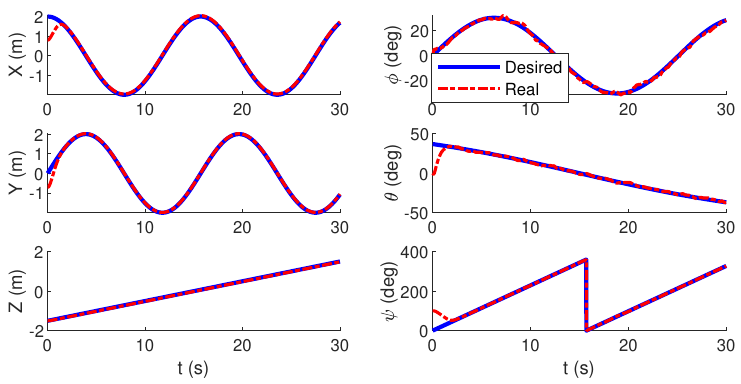}
				\caption{Position and Orientation Tracking} 
			\end{subfigure}
			\vspace{2ex}
			\begin{subfigure}{\linewidth}
				\includegraphics[width=0.9\linewidth]{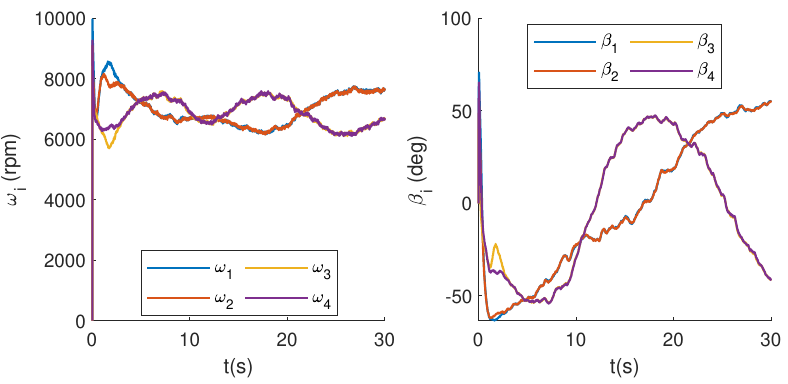}
				\caption{Control Inputs} 
			\end{subfigure}
			
			\caption{Simulation in presence of uncertainties and disturbances of the maneuver 4 for Hedral configuration using SMC.} 
			\label{fig:hedral-real-4}
\end{figure}

\subsection{Verification of Dynamic Equations} \label{subsec:verify}
The \textit{MATLAB Simscape Multibody} environment computes the system's dynamics using the physical parameters of the \ac{CAD} model, rather than relying on user-defined equations of motion. We implemented the \ac{SMC} controller, which is based on the dynamic equations derived in Section \ref{sec:dynamics} for the Hedral configuration, to control its corresponding \ac{CAD} model. 

\begin{figure}
			\begin{subfigure}{0.9\linewidth}
				\includegraphics[width=\linewidth]{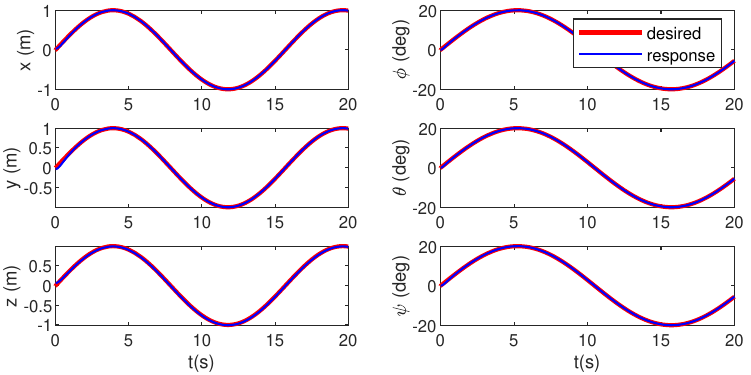}
				\caption{Position and Orientation Tracking} 
			\end{subfigure}
			\vspace{2ex}
			\begin{subfigure}{0.9\linewidth}
				\includegraphics[width=\linewidth]{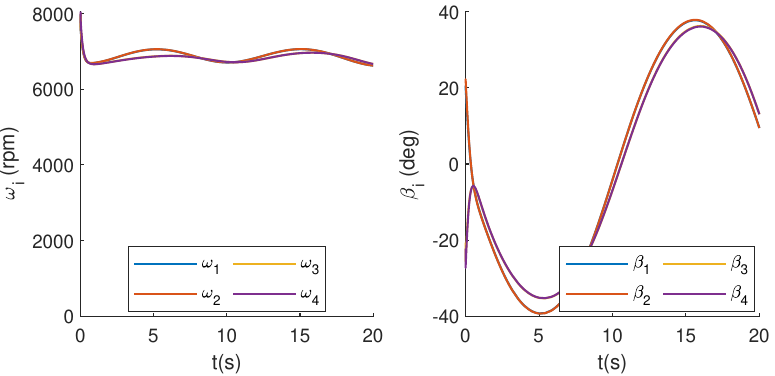}
				\caption{Control Inputs} 
			\end{subfigure}
			
			\caption{Results of SMC implementation on a CAD model in the \textit{Simscape Multibody} environment for Maneuver 3.} 
			\label{fig:verify}
		\end{figure}

\begin{figure}
			\centering
			\includegraphics[width=7.3cm]{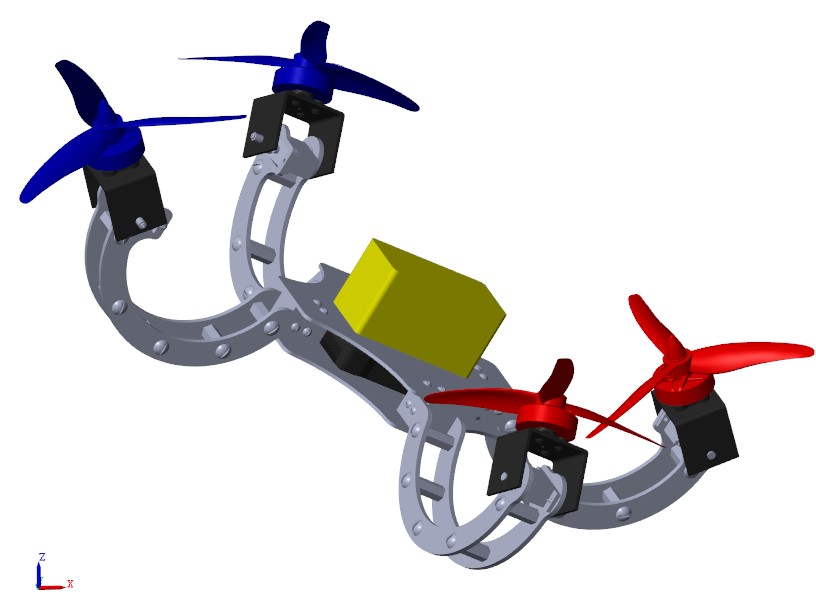}
			\caption{Snapshot from a simulation in the \textit{Simscape Multibody} environment, showing the implementation of the \ac{SMC} to control a \ac{CAD} model of Hedral configuration, for verification of dynamic equations by the methodology discussed in Subsection \ref{subsec:verify}.}
			\label{fig:simscape}
		\end{figure}

The \ac{CAD} model is designed to closely align with the assumptions made in Section~\ref{sec:dynamics}. To ensure the effectiveness of our methodology, the term  $\bm{k} \pdot \tanh{(\bm{\sigma} \pdot \bm{s})}$, which enhances the controller's robustness, was omitted. 
As depicted in Figure~\ref{fig:verify}, the controller effectively steers the system to accurately track the desired trajectory in maneuver 3, confirming the accuracy of the derived dynamic equations. Figure~\ref{fig:simscape} presents a snapshot from the simulation conducted in the \textit{Simscape} environment.		
		
		\section{Comparative Analysis and Design Roadmap}
		\subsection{Energy Considerations}
		Power consumption is a crucial concern in drones given their constrained battery capacity. For a comprehensive comparison across configurations and controllers in terms of power usage, we introduce factors. The factors neglect the low battery usage of the servo motors. The controllers are designed with the aim of minimizing the factors. 
		
		Considering the direct relationship of thrust and drag torque generated by a motor-propeller pair with the square of its rotational velocity, the Power Consumption Factor (PCF) is defined as
	\begin{equation}
		\label{PCF}
		\text{PCF}=\frac{1}{t} \int_{0}^{t} (\omega_1^4+\omega_2^4+\omega_3^4+\omega_4^4)^{1/2} \, dt ,
	\end{equation}
	where $t$ denotes the time.
	
	To have a better perspective in terms of comparison, the Nondimensionalized Power Consumption Factor (NPCF) is introduced by the following method.
	
	The static equilibrium with $\beta_i=0$ is expressed as
	\begin{equation}
		4 k_f \omega_h^2=mg.
	\end{equation}
	
	Thus,
	\begin{equation}
		\omega_h^2=\frac{mg}{4kf} .
	\end{equation}
	where $\omega_h$ is the minimum velocity of propellers to hover the robot. 
	
	\ac{NPCF} is defined as
	\begin{equation}
		\text{NPCF}=\frac{1}{t} \int_{0}^{t} \left( \frac{\omega_1^4+\omega_2^4+\omega_3^4+\omega_4^4}{4 \omega_h^4}  \right)^{1/2} \, dt .
	\end{equation}
	
	Therefore,
		\begin{equation}
		\label{NPCF}
		\text{NPCF}=\frac{2k_f}{mgt} \int_{0}^{t} (\omega_1^4+\omega_2^4+\omega_3^4+\omega_4^4)^{1/2} \, dt .
	\end{equation}
	
	Actually, NPCF defined as the average effort of the robot relative to the minimum effort needed for steady hovering. A lower NPCF value indicates less power consumption.
	
	A \ac{CF} is defined as (\ref{CF}) to compare the power usage of two items by percentage.
	\begin{equation}
		\label{CF}
		\text{CF}=\frac{2|\text{NPCF}_1-\text{NPCF}_2|}{\text{NPCF}_1+\text{NPCF}_2} \times 100 \%.
	\end{equation}

		 The ideal maneuver 3 in a long flight time is taken into account for comparison in this context because it involves motions with non-zero first and second derivatives across all degrees of freedom.

		
		{\renewcommand{\arraystretch}{1.5}
			\begin{table}[h!]
				\caption{Values of Nondimensionalized Power Consumption Factor (NPCF) in the ideal maneuver 3.}
				\centering
				\fontsize{8.5pt}{8.5pt}\selectfont
				\label{tab:power}
		\begin{tabular}{| m{1.5cm} | >{\centering\arraybackslash}m{1.3cm} | >{\centering\arraybackslash}m{1.3cm} |>{\centering\arraybackslash}m{1.4cm}|}
			\hline
			controller, \ac{CoG} eccentricity & Hedral & Tilt & Tilt-Hedral \\
			\hline
			SMC, \;\;\; e=0 & 1.0546 & 1.0546 & 0.9999 \\
			\hline
			SMC, e=0.05 m & 1.0580& 1.0580 & 1.0028 \\
			\hline
			PID-based, e=0 & 1.087 & 1.087 & 1.0428 \\
			\hline
			PID-based, e=0.05 m & 1.098 & 1.098 & 1.0477 \\
			\hline
		\end{tabular}

			\end{table}
		}
		
Here are some notable conclusions from Table \ref{tab:power} using \ac{CF}:

\begin{itemize}
	\item \textbf{Eccentricity of \ac{CoG}}: The eccentricity of the \ac{CoG} along the body $z$ direction leads to higher power usage. The Hedral configuration with \( e = 0 \) exhibits a \textbf{0.32\%} lower power consumption compared to \( e = 0.05 \) when using \ac{SMC}. This value increases to \textbf{1.01\%} with a PID-based controller, indicating that \ac{SMC} better manages this eccentricity.
	
	\item \textbf{Comparison between configurations}: \ac{PCF} for Tilt and Hedral is equal. Tilt-Hedral configuration demonstrates lower energy consumption, aligning with expectations due to its enhanced actuation setup, resulting in improved optimization potential. Using the \ac{SMC} controller for configurations with \( e = 0 \), the Tilt-Hedral exhibited \textbf{5.41\%} lower power usage. Note that this comparison does not consider the mass of the four additional servo motors in the Tilt-Hedral configuration. Additionally, the initial construction cost of this configuration is relatively higher than the others.

	\item \textbf{\ac{SMC} vs. PID-based controller}: The \ac{SMC} performs better in terms of power usage compared to the PID-based controller, especially in the presence of \ac{CoG} eccentricity. 
			The Hedral configuration with \( e = 0.05 \) shows a \textbf{3.71\%} lower power consumption when incorporating \ac{SMC} instead of a PID-based controller. This is because the allocation method used in the \ac{SMC} allows for minimizing power usage in every step of computation, whereas the PID-based controller is designed to use a constant allocation matrix to minimize computation costs. This difference in power consumption is more evident in the presence of disturbances and uncertainties.
		
\end{itemize}

		
		\begin{figure*}
			\centering
			\includegraphics[width=0.75 \linewidth]{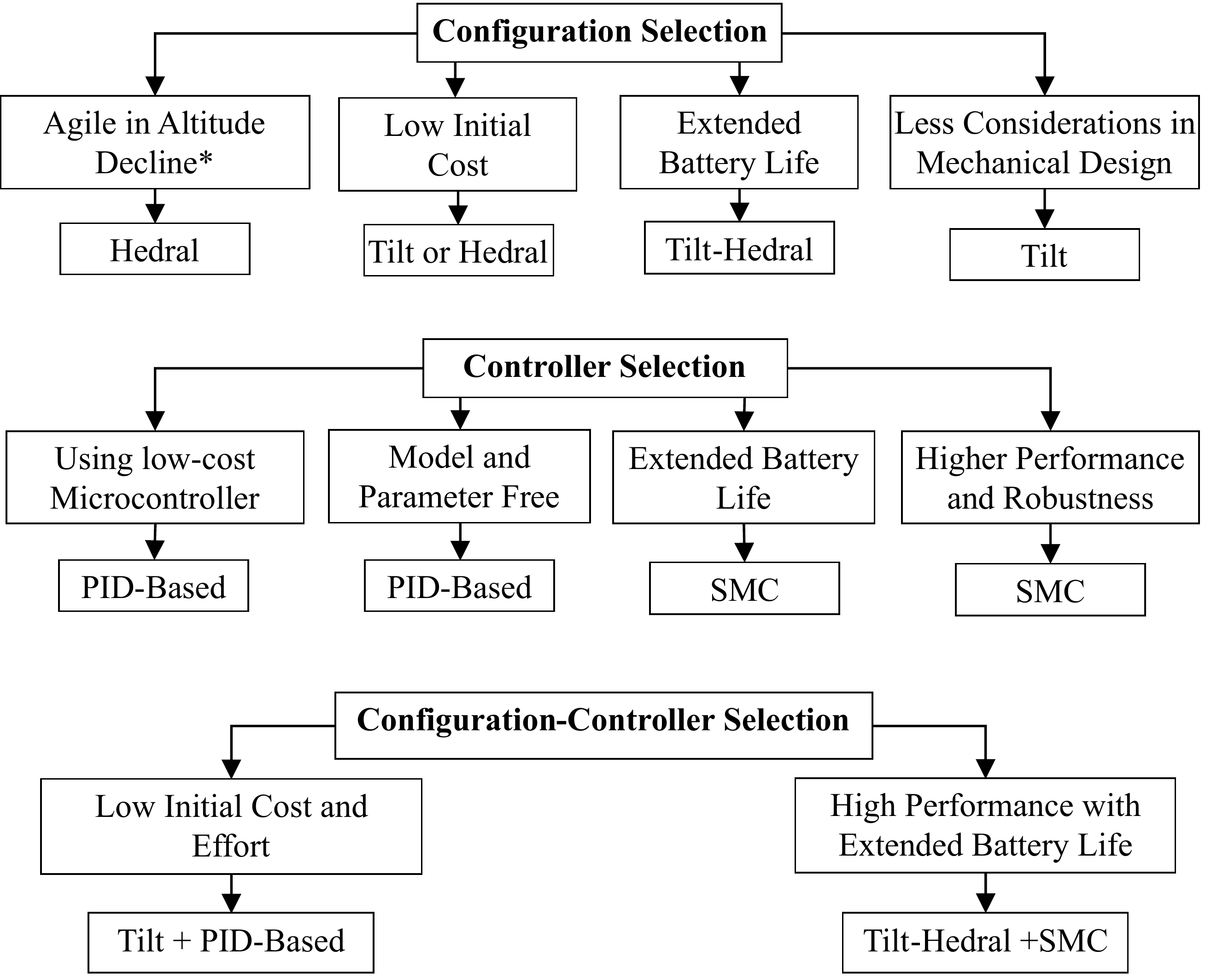}
			\caption{Selection of configuration and controller for omniorientational drone based on design objectives. \\*Agility in altitude decline by incorporating hedral angle to mitigate "vortex ring state" is discussed by \citet{tala-vortex}.}
			\label{fig:comp confs}
		\end{figure*}
		
		\begin{figure*}
			\centering
			\includegraphics[width=0.75 \linewidth]{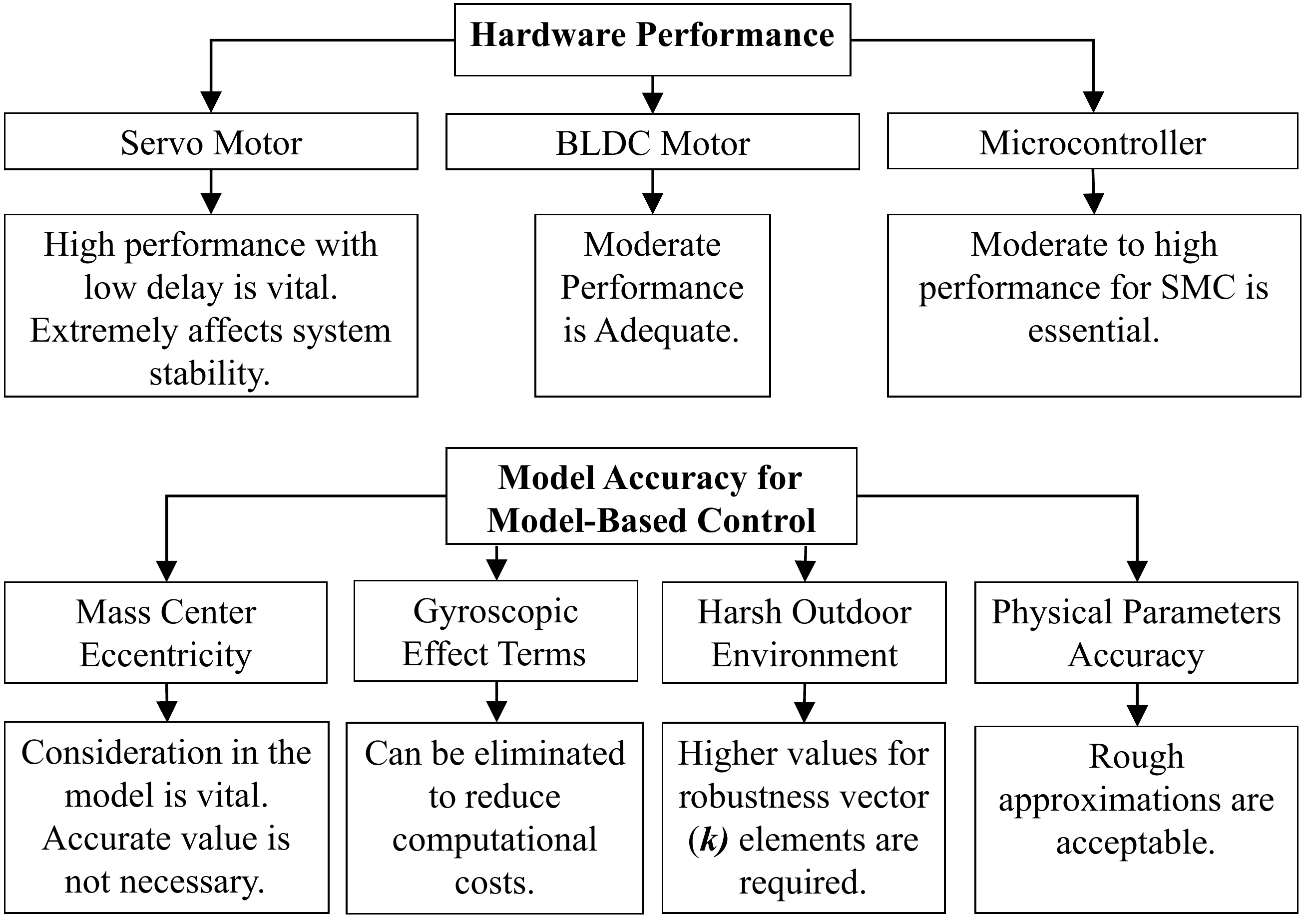}
			\caption{Summarizing the tips on required hardware performance and model accuracy.}
			\label{fig:comp model}
		\end{figure*}		
		
        \subsection{Fabrication Considerations}
        \label{subsec:fab}
		Some of the disturbances, uncertainties, and unmodeled dynamics are compared in Table \ref{tab:perturbations}. This provides valuable insights into areas that warrant further attention for model improvement and identifies areas where additional efforts may not yield significant improvements in efficiency or accuracy. 		
		\acrodef{UIF}{Uncertainty Impact Factor}
			
		\textbf{Defining the methodology}:
		The \ac{UIF} values are collected from numerous simulations. The conclusions are based on \textit{the impact of an uncertainty on the systems, resulting in an increase of \ac{SMC} robustness gains (see Equation (\ref{sdot-tanh})) to compensate} \textemdash achieving a system response that closely matches the original behavior in terms of stability, settling time, and tracking error, despite the presence of disturbances or uncertainties. Therefore, 
		\begin{equation}
			UIF=\frac{max(\bm{k}_n)}{max(\bm{k}_o)},
		\end{equation}
		where $max(\bm{k}_n)$ and $max(\bm{k}_o)$  represent the maximum element of the gain vector $\bm{k}$ in the presence and absence of a specific disturbance or uncertainty, respectively.
		The values of the uncertainties for this study are in the range defined in \ref{subsec:realsim}.
		
		 According to Table \ref{tab:perturbations}, some tips can be deducted for real-world scenarios:
		
		{\renewcommand{\arraystretch}{1.5}
			\begin{table}
				\centering
				\fontsize{8pt}{8pt}\selectfont
				\caption{Effects of various types of uncertainties in tracking and stability of the control system. Note: The uncertainties for this study are in the range defined in Subsection~\ref{subsec:realsim}.}
				\label{tab:perturbations}
				\begin{tabular}{| >{\centering\arraybackslash}m{2.7cm} | >{\centering\arraybackslash}m{0.8cm}| >{\centering\arraybackslash}m{1.3cm}|>{\centering\arraybackslash}m{0.9cm}| }
					\hline
					Type of Uncertainty & UIF & Comparative Effect & Design Considerations \\
					\hline
					random disturbance forces& 5 to 7 & high & \ref{subsec:fab}-a\\
					\hline
					constant disturbance forces& $\approx$2& intermediate   & \ref{subsec:fab}-a \\
					\hline
					error of servo motors& $\approx$3 & intermediate & \textemdash \\
					\hline
					error of BLDC motors&  $\approx$2.2 & intermediate  & \textemdash \\
					\hline
					delay of servo motors& $\approx$2.5& intermediate & \ref{subsec:fab}-b \\
					\hline
					delay of BLDC motors & $\approx$1.7 &low &  \textemdash \\
					\hline
					uncertain mass  & $\approx$1.2 & very low & \ref{subsec:fab}-c  \\
					\hline
					uncertain aerodynamic coefficients of propellers & $\approx$1.5 & low & \ref{subsec:fab}-c \\
					\hline
					uncertain inertia moments  & $\approx$1.9 & low & \ref{subsec:fab}-c  \\
					\hline
					unmodeled gyroscopic effect & $\approx$1.5 & low & \ref{subsec:fab}-d  \\
					\hline
					unmodeled CoG eccentricity & 20 to 30 & very high & \ref{subsec:fab}-e \\
					\hline
				\end{tabular}
			\end{table}
		}

		\textbf{a)} Designing the drone for harsh outdoor environments affected by wind or other disturbances necessitates the use of higher gains, resulting in a need for more costly actuators. However, if a constant wind disturbance\textemdash such as indoor ventilation\textemdash is present, the design requires fewer considerations.
		
		\textbf{b)} Delay of servo motors is dangerous because it leads to instability and failure of the system\textemdash not only tracking errors. So, using cheap servo motors would increase the risk of crash.
		
		\textbf{c)} It is not necessary to conduct highly precise measurements to determine mass and moments of inertia. Values obtained from a CAD software for a moderately detailed assembly design would be acceptable for the designed controller. Similarly, for estimating the aerodynamic coefficients of the propeller, approximate calculations based on motor catalog data for a similar propeller are adequate, eliminating the need for precise testing with a motor-propeller tester. 
		
		\textbf{d)} The gyroscopic effect can be disregarded in the controller model to decrease the load of real-time computations by the microprocessor.
		
		\textbf{e)} Neglecting eccentricity of \ac{CoG} within the model used by the \ac{SMC} significantly challenges trajectory tracking. In such case, \ac{SMC} necessitates control gains that are 20 to 30 times higher compared to instances where eccentricity is included in the controller's model. This elevated gain requirement results in notable chattering and overshoot of control signals, potentially leading to actuator saturation. When the eccentricity of \ac{CoG} is modeled in the controller, even with a non-accurate value, the controller can easily compensate for the error. Also, it is advisable to design the drone in a way that minimizes this eccentricity.
		
		\subsection{Overall Comparisons and Design Roadmap}
		Tables \ref{tab:overall} and \ref{tab:overall controller} provide an overall comparison of the configurations and proposed controllers, highlighting their advantages and limitations under the same conditions.
		
		{\renewcommand{\arraystretch}{1.5}
			\begin{table}[h]
				\caption{Overall comparison of the configurations.}
				\centering
				\fontsize{9pt}{9pt}\selectfont
				\label{tab:overall}
				\begin{tabular}{| >{\centering\arraybackslash}m{1.8cm} | >{\centering\arraybackslash}m{4.8cm}| }
					\hline
					Initial Cost & Tilt-Hedral> Hedral  $\approx$ Tilt \\
					\hline
					Energy Consumption & Hedral $\approx$ Tilt > Tilt-Hedral \\
					\hline
					Physical Limitations & Hedral configuration has more limitations in manipulating the propeller axes, demanding more considerations in mechanical design, compared to Tilt configuration.  \\
					\hline
				\end{tabular}
			\end{table}
		}
		
		{\renewcommand{\arraystretch}{1.5}
			\begin{table}[h]
				\caption{Overall comparison of the proposed controllers.}
				\fontsize{9pt}{9pt}\selectfont
				\centering
				\label{tab:overall controller}
				\begin{tabular}{| >{\centering\arraybackslash}m{3.5cm} | >{\centering\arraybackslash}m{3cm}| }
					\hline
					Power Consumption & PID-based > SMC \\
					\hline
					Disturbance Rejection & SMC > PID-based \\
					\hline
					Real-Time Computation Cost & SMC > PID-based\\
					\hline
				\end{tabular}
			\end{table}
		}
		
Charts in Figures \ref{fig:comp confs} and \ref{fig:comp model} summarize the discussed analysis in this section, providing a roadmap and practical insights in designing omniorientational drones.

		\section{Conclusion}
This paper presents several distinct omniorientational drone configurations aimed at overcoming the limitations of conventional multi-rotors by enabling full independent control across all six degrees of freedom. The dynamic equations for each configuration were derived, and two control strategies were designed based on sliding mode and PID. The validity of these equations was confirmed by applying the sliding mode controller to a CAD model. While one configuration fell short of meeting the objectives, the remaining three demonstrated robust performance in simulations, successfully executing maneuvers beyond the capabilities of conventional multi-rotors under harsh disturbances and uncertainties.
To compare the performance of different drone configurations and controllers, the Power Consumption Factor (PCF) was introduced. This metric was also utilized to assess the impact of parameter uncertainties, disturbances, and center-of-gravity eccentricity. Furthermore, a control allocation criterion was implemented to optimize battery usage by minimizing the PCF.
Finally, this work provides a roadmap for future researchers to design omniorientational drones tailored to specific objectives, offering practical insights into configuration selection and controller design. The authors' future goal is to leverage the outcomes of this study to fabricate an optimally designed drone and validate the results through experimental testing.

		\section*{Declaration of Generative AI and AI-assisted technologies in the writing process}
		During the preparation of this work, the authors used ChatGPT, an AI language model, to refine statements, improve readability, enhance grammar, and check for spelling. After using this tool, the authors reviewed and edited the content as needed and take full responsibility for the content of the publication.


\section*{Funding Sources}
This research did not receive any specific grant from funding agencies in the public, commercial, or not-for-profit sectors.

\section*{Deceleration of Interests}
The authors declare that they have no known competing financial interests or personal relationships that could have appeared to influence the work reported in this paper.

		\bibliographystyle{elsarticle-num-names} 
		\bibliography{Ali_bib}
		
		
		
		
		
	\end{document}